%% file: main.tex
\documentclass{article} 
\usepackage{iclr_preprint,times}

\input{math_commands.tex}

\usepackage{hyperref}
\usepackage{url}

\usepackage{microtype}
\usepackage{graphicx}    
\usepackage{booktabs}    
\usepackage{amsmath,amsfonts,bm,amsthm,amssymb}
\usepackage{hyperref}
\usepackage{url}
\usepackage{enumitem}
\usepackage{listings}
\usepackage{xcolor}

\definecolor{codegreen}{rgb}{0,0.6,0}
\definecolor{codegray}{rgb}{0.5,0.5,0.5}
\definecolor{codepurple}{rgb}{0.58,0,0.82}
\definecolor{backcolour}{rgb}{0.95,0.95,0.92}

\lstdefinestyle{mystyle}{
    backgroundcolor=\color{backcolour},   
    commentstyle=\color{codegreen},
    keywordstyle=\color{magenta},
    numberstyle=\tiny\color{codegray},
    stringstyle=\color{codepurple},
    basicstyle=\ttfamily\scriptsize,
    breakatwhitespace=false,         
    breaklines=true,                 
    captionpos=b,                    
    keepspaces=true,                 
    numbers=left,                    
    numbersep=5pt,                  
    showspaces=false,                
    showstringspaces=false,
    showtabs=false,                  
    tabsize=2
}

\newtheorem{theorem}{Theorem}
\newtheorem{proposition}[theorem]{Proposition}
\newtheorem{lemma}[theorem]{Lemma}
\newtheorem{corollary}[theorem]{Corollary}
\theoremstyle{remark}
\newtheorem{remark}[theorem]{Remark}
\theoremstyle{definition}
\newtheorem{definition}[theorem]{Definition}

\newcommand{\norm}[1]{\|#1\|}
\newcommand{\supp}{\operatorname{supp}}
\def\imp{\includegraphics[width=1em,trim=0 .4em 0 0]{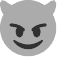}}

\title{Stochastic non-metric MDS\\for noisy high-dimensional data}

\author{Sebastian Damrich\\
University of T\"{u}bingen \\
\texttt{sebastian.damrich@uni-tuebingen.de} \\
\And
Dmitry Kobak \\
Ghent University and VIB.AI\\
\texttt{dmitry.kobak@ugent.be}}

\iclrfinalcopy 
\begin{document}

\maketitle

\begin{abstract}
Multidimensional scaling (MDS) constructs low-dimensional embeddings that approximate pairwise distances in a high-dimensional dataset. MDS exists in several flavors, with metric MDS approximating distances directly and non-metric MDS additionally optimizing for an arbitrary monotonic transformation of the high-dimensional distances. 
Empirically, metric MDS is known to suffer from the curse of dimensionality. Here we extend prior theoretical results for metric MDS on fully uninformative data to a more general case of finite, high-dimensional data with concentrated distances. We show that the metric MDS embeddings of such datasets stay distributionally close to the data-independent limit case. We then show that non-metric MDS is more robust to high-dimensional noise and can yield informative embeddings of such data. 
While existing implementations of non-metric MDS are slow, we develop a fast implementation using stochastic gradient descent in PyTorch. Using our implementation, we conduct an empirical study using multiple simulated and real-world datasets, including population-genomic and scRNA-seq data, and show that non-metric MDS can strongly outperform other methods in terms of global structure preservation.
\end{abstract}

\section{Introduction}

Multidimensional scaling (MDS) \citep{borg2005modern} is one of the classic dimensionality-reduction and embedding methods developed in the 1960s \citep{shepard1962analysis1,shepard1962analysis2,kruskal1964multidimensional,kruskal1964nonmetric}. Its goal is to find low-dimensional embedding vectors $\mathbf z_i \in \mathbb R^p$ such that pairwise distances $\|\mathbf z_i - \mathbf z_j\|$ between them approximate some given pairwise distances $D_{ij}$. MDS is well-known in the field of dimensionality reduction as the algorithm aiming to preserve global structure of the data \citep{lee2007nonlinear}, as opposed to various neighbor-embedding algorithms such as $t$-SNE \citep{maaten2008visualizing} and UMAP \citep{mcinnes2018umap} that aim to preserve local structure.

Empirically, MDS often struggles to produce a meaningful embedding of high-dimensional data \citep{bodt2025lowdim}. Here we formalize this observation and study the behavior of MDS under distance concentration. We show that the so-called \textit{metric MDS}, where $\|\mathbf z_i - \mathbf z_j\|$ should directly approximate given $D_{ij}$, suffers from the curse of dimensionality and remains close to a data-independent embedding~\citep{mattner1992extremal, buja1994inequalities}. In contrast, we demonstrate that \textit{non-metric MDS}, a more flexible but less widely used flavor of MDS, where $\|\mathbf z_i - \mathbf z_j\|$ should approximate $D_{ij}$ only up to an arbitrary monotonic transformation (i.e. approximating the order of distances and not the distances themselves), is more robust and can better tolerate high-dimensional noise, while still producing informative embeddings.

Most existing implementations of non-metric MDS, such as the one in scikit-learn \citep{pedregosa2011scikit} in Python or the \texttt{smacof} package in R \citep{leeuw2009multidimensional,mair2022more}, have quadratic memory and cubic runtime complexity, precluding its use for datasets with sample sizes $n\gg 10\,000$. Indeed, simply storing the $n\times n$ matrix of pairwise distances $\mathbf D$ in memory can already be prohibitive for large $n$. While several stochastic versions of metric MDS have been developed in various contexts  \citep{agrafiotis2003stochastic,zheng2018graph,lambert2022squadmds,fang2025stablemds, hangan2026bridging}, fast implementations of non-metric MDS do not exist. Here we fill this gap.

In summary, our contributions are:
\begin{enumerate} 
    \item we establish a perturbation bound for the limiting solution of metric MDS;
    \item we show that non-metric MDS is more robust and can tolerate higher-dimensional noise;
    \item we develop a stochastic implementation of non-metric MDS in PyTorch;
    \item we show that non-metric MDS can outperform other methods in global embedding quality.
\end{enumerate}
We argue that our results are relevant for modern embedding methods, as many of them include MDS-based loss terms to ensure global structure preservation (see Discussion).

\section{Related Work}
\label{sec:related}

On the theoretical side, so-called classical MDS (see below) has recently received attention from mathematicians and was shown to exhibit surprising properties in the continuous $n\to\infty$ limit \citep{kroshnin2022infinite,lim2024classical}. \citet{buja1994inequalities} characterized the solution of metric MDS in the continuous limit for fully uninformative inputs.

On the practical side, multiple papers independently re-introduced stochastic \emph{metric} MDS in various contexts. \citet{agrafiotis2003stochastic} developed a stochastic metric MDS with batch size $b=2$, calling it stochastic proximity embedding (SPE). This method was developed in the context of computational chemistry, to visualize large collections of chemical compounds \citep{agrafiotis2010stochastic}.
\citet{zheng2018graph} developed stochastic implementations of weighted MDS in the context of graph drawing. In that setting, the input matrix of distances is the matrix of shortest-path graph distances between pairs of graph nodes. For optimization, the authors sampled pairs of points, essentially using mini-batches of size $b=2$. \citet{hangan2026bridging} adopted this approach for dimensionality reduction and \citet{fang2025stablemds} used a similar approach with $b=n/10$ for their FastMDS.
\citet{lambert2022squadmds} developed SQuadMDS (stochastic quartet MDS) that optimizes metric MDS using mini-batches of size $b=4$ (quartets). The authors developed a C++ implementation and used it to obtain metric MDS embeddings of MNIST and the Tasic datasets, also analyzed here. Furthermore, the authors suggested a hybrid between SQuadMDS and $t$-SNE.

At the same time, we are not aware of any scalable implementations of \textit{non-metric} MDS. All examples in \citet{borg2005modern} and \citet{mair2022more} have $n<100$.

Non-metric MDS is closely related to ordinal embeddings, which learn an embedding from a set of triplet comparisons such as ``$j$ is closer to $i$ than to $k$''. There are several landmark-based methods that provide approximate implementations suitable for large sample sizes \citep{anderton2019scaling, ghosh2019landmark}. Non-metric MDS also uses only the rank order of the high-dimensional distances, but over all pairs, and not just over a subset of triplets. Therefore we do not include ordinal embeddings into our benchmark comparison.

\section{Background}
\label{sec:background}

\paragraph{Notation} 

Let the high-dimensional dataset consist of $n$ vectors $\mathbf x_i \in \mathbb R^d$, forming an $n\times d$ data matrix $\mathbf X$. Let $\mathbf D$  be an $n\times n$ pairwise distance matrix with $D_{ij}$ being some distance (e.g. Euclidean, cosine, etc.) between $\mathbf x_i$ and $\mathbf x_j$. Alternatively, we may be given (e.g. experimentally obtain) some distance matrix $\mathbf D$ directly, without access to any underlying $\mathbf X$, and without knowing the value of $d$.

The goal of any MDS algorithm is to construct low-dimensional embedding vectors $\mathbf z_i \in \mathbb R^p$ such that their pairwise distances approximate pairwise distances $D_{ij}$. Typically $p=2$ for visualization purposes. We refer the reader to \citet{borg2005modern} for a detailed systematic treatment of MDS.

\paragraph{Classical MDS}

In classical MDS, also known as Torgerson's scaling \citep{torgerson1952multidimensional} and as principal coordinates analysis (PCoA) \citep{gower1966some}, the assumption is that $\mathbf D$ is a matrix of Euclidean distances. If so, it can be converted into the Gram matrix of centered vectors $G_{ij} = (\mathbf x_i - \bar{\mathbf x})^\top (\mathbf x_j - \bar{\mathbf x})$ by double-centering the matrix of squared distances:
\begin{equation}
    \mathbf G = -(\mathbf I - \mathbf 1_n/n)(\mathbf D \odot \mathbf D / 2)(\mathbf I - \mathbf 1_n/n),
\end{equation}
where $\odot$ denotes elementwise multiplication, and $\mathbf 1_n$ denotes the $n\times n$ matrix of all ones. If $\mathbf G = \mathbf U \boldsymbol \Lambda \mathbf U^\top$ is the eigendecomposition of $\mathbf G$, then $\mathbf Z = \mathbf U \boldsymbol \Lambda^{1/2}$ gives the principal components of $\mathbf X$ and the classical MDS of $\mathbf D$.

Note that if $\mathbf D$ is not a matrix of Euclidean distances, then classical MDS is not equivalent to principal component analysis (PCA) of $\mathbf X$. If $\boldsymbol \Lambda$ contains negative eigenvalues, then $\mathbf D$ cannot be obtained as Euclidean distances between vectors of any dimensionality \citep{schoenberg1935remarks}. Recent work studied properties and extensions of classical MDS in this regime \citep{sonthalia2021how,deng2024neucmds}, but this is very different from non-metric MDS which learns an ordinal transformation while retaining Euclidean output metric.

\paragraph{Metric MDS}

The loss function of metric MDS is the squared error between $D_{ij}$ and $\|\mathbf z_i - \mathbf z_j\|$:
\begin{equation}
    \mathcal L_\text{mMDS} = \sum_{i<j} \big(\|\mathbf z_i - \mathbf z_j\| - D_{ij}\big)^2.
    \label{eq:stress}
\end{equation}
It is called (raw) stress. It is often more convenient to work with the raw stress divided by $n(n-1)/2$, which we will call the average stress. The common approach for minimizing the stress loss function relies on the iterative SMACOF algorithm (scaling by majorizing a complicated function; \citealp{leeuw1977application}), which in this case is equivalent to gradient descent with learning rate $1/(2n)$. This algorithm is used in most common MDS implementations, e.g. in R \citep{leeuw2009multidimensional,mair2022more} and in the scikit-learn library \citep{pedregosa2011scikit} in Python. SMACOF converges to a local minimum of $\mathcal L_\text{mMDS}$, but note that $\mathcal L_\text{mMDS}$ is not convex and has many local minima, so the final embedding depends on the initialization.

\paragraph{Non-metric MDS}

Non-metric MDS \citep{shepard1962analysis1,shepard1962analysis2,kruskal1964multidimensional,kruskal1964nonmetric} additionally allows an arbitrary monotonic transformation $f(\cdot)$ of the input distances:
\begin{equation}
    \mathcal L_\text{nMDS} = \sum_{i<j} \big(\|\mathbf z_i - \mathbf z_j\| - f(D_{ij})\big)^2 \quad\text{s.t.}\quad \frac{2}{n(n-1)}\sum_{i<j}f(D_{ij})^2 = 1.
    \label{eq:nmds}
\end{equation}
Here, to avoid a trivial solution with $f(\cdot) = 0$, the transformed values $f(D_{ij})$ are usually constrained to have mean squared off-diagonal value equal to 1 \citep{leeuw2009multidimensional,mair2022more}. Fixing $f(\cdot)$ to be proportional to the identity recovers metric MDS up to the scaling factor. The standard approach for minimizing this loss function is alternating minimization: after each step of the SMACOF algorithm, an optimal $f(\cdot)$ is fit by isotonic (monotonic) regression of the current $\|\mathbf z_i - \mathbf z_j\|$ values on $D_{ij}$ values. The resulting $f(D_{ij})$ values are then scaled to satisfy the constraint.

\paragraph{Stress-1}

It is often more convenient to use some normalized version of stress, for example so-called \mbox{stress-1}~\citep{kruskal1964multidimensional,kruskal1964nonmetric}:
\begin{equation}
    \text{Stress-1} = \sqrt{\frac{\sum_{i<j} \big(\|\mathbf z_i - \mathbf z_j\| - f(D_{ij})\big)^2}{\sum_{i<j} \|\mathbf z_i - \mathbf z_j\|^2}},
    \label{eq:stress1}
\end{equation}
where $f$ is the identity in the case of metric MDS. For fixed $D_{ij}$ and $f$, stress-1 depends on the overall scale of $\mathbf Z$. A minimizer of the raw stress is, up to a scale factor, also a minimizer of stress-1. We therefore always report the stress-1 after rescaling $\mathbf Z$ by the optimal factor. The optimal factor can be computed analytically: with $\mathbf d = f(D_{ij})$ and $\mathbf d' = \|\mathbf z_i - \mathbf z_j\|$, both flattened, it is $\|\mathbf d\|^2 / \langle \mathbf d, \mathbf d' \rangle$, and the resulting stress-1 is $\sqrt{1 - \cos^2\angle(\mathbf d, \mathbf d')}$.

If $\mathbf Z$ is a minimizer of raw stress (Eq.~\ref{eq:stress} or Eq.~\ref{eq:nmds}), the stress-1 of the optimally rescaled $\mathbf Z$ can also be obtained from the stress-1 $s$ of $\mathbf Z$ itself (note that scikit-learn returns $s$): it equals $\sqrt{s^2 / (1 + s^2)}$, and the optimal factor is $1 + s^2$ \citep[Chapter 11.1]{borg2005modern}.

Furthermore, at such a minimum, the rescaled stress-1 equals the square root of the raw stress divided by $\sum_{i<j} f(D_{ij})^2$, the squared norm of the target distances \citep[Chapter 11.1]{borg2005modern}. If $f(\cdot)$ is normalized as in Eq.~\ref{eq:nmds}, this squared norm is simply the number of pairs, $n(n-1)/2$, so the rescaled stress-1 coincides with the square root of the average stress, i.e. of the loss that is being minimized.

\paragraph{Scikit-learn implementation and our improvements}

The scikit-learn library \citep{pedregosa2011scikit} in Python contains implementations of metric and non-metric MDS using the SMACOF algorithm. While working on this paper, we identified a number of issues with that implementation and fixed them in a series of four PRs (PR numbers hidden to preserve anonymity), which have been merged into the version 1.8 (released in December 2025). Apart from fixing some implementation errors in non-metric MDS, we implemented the classical MDS algorithm (as \texttt{manifold.ClassicalMDS} class) and set it as default initialization for metric and non-metric MDS. All scikit-learn experiments reported below used version 1.8 and default hyperparameters (default from version 1.10), and the algorithm always reached the default convergence criterion.

\section{The curse of dimensionality for metric MDS}
\label{sec:curse}

\begin{figure}[t]
\begin{minipage}[c]{0.65\textwidth}
\includegraphics{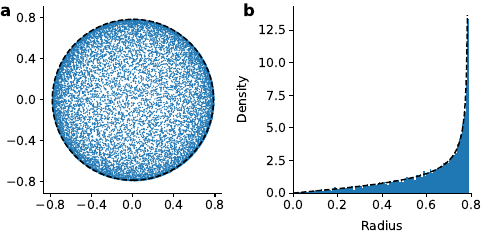}
\end{minipage}\hfill
\begin{minipage}[c]{0.35\textwidth}
\vspace{-2em}
\caption{\textbf{(a)} Metric MDS of a $10\,000\times10\,000$ distance matrix with all off-diagonal elements equal to $1+\varepsilon_{ij}$ with $\varepsilon_{ij}\sim\mathcal{N}(0, 0.01)$. 
The dashed circle has radius $\pi/4$. \textbf{(b)} Radial density of points in panel (a). Linear increase corresponds to the uniform 2D density. The dashed line shows theoretical prediction.}
\label{fig:limit}
\end{minipage}
\end{figure}

Here we will extend and combine prior results by~\citet{mattner1992extremal}, \citet{buja1994inequalities}, and \citet{aggarwal2001surprising} to show that finite, noisy, high-dimensional data yields an uninformative metric MDS embedding. The motivates the use of non-metric MDS for high-dimensional data. 

The standard result on distance concentration is that in the presence of high-dimensional noise, all pairwise distances become approximately the same~\citep{aggarwal2001surprising}. For instance, consider an arbitrary 2D dataset with $n$ points (e.g. sampled from a mixture of well-separated Gaussians, as we will do in Section~\ref{sec:algorithm}). To mimic high-dimensional noise, we append $d$ additional non-discriminative dimensions with i.i.d. values with mean 0, variance $\sigma^2$, and finite 4th moment, so that the resulting dataset has dimensionality $d+2$. As $d$ increases, all pairwise distances $D_{ij}$ will concentrate around $\sqrt{2d\sigma^2}$ as all vectors become close to orthogonal and each of them has squared norm close to $d\sigma^2$~\citep{hall2005geometric,hiraoka2026curse}.

Dividing all distances by $\sqrt{2d\sigma^2}$, we see that the $n\times n$ distance matrix approaches $D_{ij} = 1 - \delta_{ij}$ with all off-diagonal elements equal to 1. 
The continuous limit of metric MDS, corresponding to $n\to \infty$, on this fully uninformative input was analyzed by~\citet{mattner1992extremal} and \citet{buja1994inequalities}. Let $\mathcal{P}(\R^2)$ denote the set of probability measures on $\R^2$. For $\rho \in \mathcal{P}(\R^2)$ and $X, Y \overset{\text{i.i.d.}}{\sim} \rho$ the average stress functional is
\begin{equation}
  \label{eq:stress-functional-maintext}
  S[\rho]
  =
  \E_\rho\!\bigl[(\norm{X-Y}-1)^2\bigr] =  \iint \big(\|x-y\|-1\big)^2 \rho(x) \rho(y) \, dx \, dy.
\end{equation}
\begin{theorem}[Limiting metric MDS embedding, \citep{mattner1992extremal, buja1994inequalities}]
  \label{thm:main}
  The unique minimizer of $S[\rho]$ over $\mathcal{P}(\R^2)$ with $\mathbb E[\rho] = 0$ is the Riesz $s=1$ equilibrium measure with density
  \begin{equation}
    \rho^*(x)
    =
    \frac{2}{\pi^2\,\sqrt{(\pi/4)^2 - \norm{x}^2}} \text{ with support on } \|x\| < \frac{\pi}{4}.
  \end{equation}
  It achieves an average stress value of $S^*:=S[\rho^*]= 1 - \frac{\pi^2}{12} \approx 0.1775$.
\end{theorem}
We include a detailed proof of this theorem in Appendix~\ref{sec:proofs}. 
The main observation is that the Laplacian of the functional derivative of $S$ yields the Riesz $s=1$ potential. This is solved by the Riesz $s=1$ equilibrium measure~\citep{riesz1938integrales, landkof1972foundations, chafai2022threshold}. 
Different from the idealized setting, real data is finite and pairwise distances are only concentrated, not exactly equal. Using the energy distance $D^2_E$~\citep{szekely2013energy}, defined for two (possibly discrete) distributions as
  \begin{equation}
    \label{eq:energy_dist_main}
    D_E^2(\mu,\nu)
    =
    2\iint \norm{x-y}\,d\mu(x)\,d\nu(y)
    - \iint \norm{x-x'}\,d\mu(x)\,d\mu(x')
    - \iint \norm{y-y'}\,d\nu(y)\,d\nu(y'),
  \end{equation}
we establish the following perturbation bound for this setting:
\begin{proposition}
\label{prop:perturbation_bound}
    Let $D'_{ij} = 1 - \delta_{ij}+ \varepsilon_{ij}$ be an $n\times n$, zero-diagonal, symmetric dissimilarity matrix, so that its root-mean-squared deviation from the fully uninformative $D$ is $\Delta = \sqrt{\frac{2}{n(n-1)}\sum_{i<j} \varepsilon_{ij}^2}$. Then for any global minimizer $X'$ of metric MDS for dissimilarities $D'$, we have
    \begin{equation}
            D_E^2(X',\rho^*)
    \;\leq\;
    \frac{\pi^2}{24n}
    \;+\; 2\sqrt{S^*}\,\Delta
    \;+\; 2\,\Delta^2
    \;\approx\;
    \frac{0.411}{n} + 0.843\,\Delta + 2\,\Delta^2.
    \end{equation}
\end{proposition}
Appendix~\ref{app:perturbation} contains the proof, which relies on bounding the energy distance by the stress functional, taking finite samples from $\rho^*$, and using a reverse triangle inequality. 

The marginal distribution of the norm $\norm{X}$ for $X \sim \rho^*$ is a rescaled Kumaraswamy $(a=2,\,b=1/2)$ distribution on $[0,\pi/4)$ with density $p(r) = 4r / \big(\pi \sqrt{(\pi/4)^2 - r^2}\big)$ (Corollary~\ref{cor:norms}). Our simulation with $n=10\,000$ (Figure~\ref{fig:limit}) nearly equidistant points 
confirms that the optimum of metric MDS stays close to the continuous limit.

\noindent\fbox{%
\parbox{\dimexpr%
    \linewidth-2\fboxsep-2\fboxrule}%
    {\imp{} \textbf{Curse of dimensionality:} Metric MDS of noisy high-dimensional data with large sample size will look like a disk with points concentrated on the boundary, with radius $\bar D\pi/4$, where $\bar D$ is the average high-dimensional pairwise distance in the data.}}


\section{Non-metric MDS is more robust to high-dimensional noise}
\label{sec:robustness}

\begin{figure}[t]
\begin{minipage}[c]{0.62\textwidth}
\includegraphics{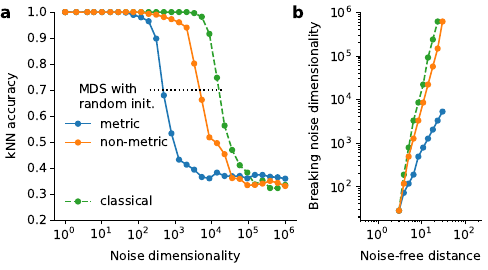}
\end{minipage}\hfill
\begin{minipage}[c]{0.38\textwidth}
\vspace{-1em}
\caption{\textbf{(a)} Leave-one-out $k$NN classification accuracy ($k=10$) in MDS embeddings of noisy three-class dataset. There are $n=99$ points from 3 equisized classes, equidistant from each other with distance $D=10$ in $\mathbb R^2$. We add $d \in [1, 10^6]$ noise dimensions with i.i.d. $\mathcal N(0, 1)$ values. Metric and non-metric MDS are run with random initializations. Average over 10 runs. \textbf{(b)} Minimum $d$ that yields the average $k$NN accuracy below $0.7$ for $D\in[3, 30]$.}
\label{fig:robustness}
\end{minipage}
\end{figure}

\definecolor{tab10orange}{RGB}{255,127,14}
\definecolor{tab10blue}{RGB}{31,119,180}
\setlength{\fboxrule}{1pt}
\setlength{\fboxsep}{2pt}
\newcommand{\mybox}[2]{{\color{#1}\fbox{\normalcolor#2}}}

We experimentally measured the robustness of metric and non-metric MDS to high-dimensional noise using the following simulation. We simulated a noisy dataset by generating $n=99$ points from 3 equisized classes, equidistant from each other with distance $D=10$ in $\mathbb R^2$. We then added $d \in [1, 10^6]$ noise dimensions with i.i.d. $\mathcal N(0, \sigma^2)$ values with $\sigma=1$. Then we ran both versions of MDS using random initialization (in order to isolate the behavior of the optimization from the quality of the initial embedding; we only used random initialization in this section) and measured the $k$NN classification accuracy in the embedding, averaged over 10 runs with different noise and random initialization. We found that non-metric MDS could tolerate much higher noise dimensionality $d$ before $k$NN accuracy collapsed (Figure~\ref{fig:robustness}a).

We then asked what value of $d$ leads to the accuracy collapse. The pairwise distances between noise vectors concentrate around $\sqrt{2d\sigma^2}$. Metric MDS considers absolute distances and gets strongly distorted when these distances become comparable to $D$. This happens when \mybox{tab10blue}{$d^*_\text{mMDS} = \Theta(D^2 / \sigma^2)$}.

For large $d$, the pairwise distances between noisy observations can be approximated by $\sqrt{2d\sigma^2 + D^2} \approx \sqrt{2d\sigma^2} + D^2 / (2\sqrt{2d\sigma^2})$. Non-metric MDS gets strongly distorted when the last term (`signal') becomes comparable to the fluctuations of the first term (`noise'), so that the noise affects the \textit{ranking} of distances. The distances between noise vectors follow a $\chi$-distribution and its standard deviation asymptotically equals $\sigma$. Equating the second term above to $\sigma$ yields \mybox{tab10orange}{$d^*_\text{nMDS} = \Theta(D^4 / \sigma^4)$}.

To verify this experimentally, we ran our simulation for various values of $D\in[3, 30]$ and found the minimal $d^*(D)$ value with the average $k$NN accuracy below 0.7 (`collapse'). In log-log coordinates, $d^*(D)$ was close to a linear function (Figure~\ref{fig:robustness}b) with regression slopes 2.2 for metric MDS and 4.1 for non-metric MDS, in an excellent agreement with our prediction. In a more varied setup, with varying clusters sizes, noise types, and only approximate equidistance, the regression slopes were all within $[2.0, 2.3]$ and $[3.9, 4.1]$, for metric and non-metric MDS, respectively (Appendix~\ref{sec:extended_robustness}).

For reference, classical MDS (equivalently, PCA) was similarly robust (Figure~\ref{fig:robustness}); random matrix theory suggests that the scaling for classical MDS is also $\Theta(D^4 / \sigma^4)$ \citep{baik2005phase}. Note that in the experiments in this section we used random initializations for metric and non-metric MDS, which is why PCA appears to perform better. In practice, we always initialize MDS with classical MDS, and below we will show that non-metric MDS can outperform PCA on real-world data. The main lesson of this section is that non-metric MDS is more robust to high-dimensional noise than metric MDS.

\section{Stochastic non-metric MDS}
\label{sec:algorithm}

\paragraph{SGD implementation in PyTorch}

To be able to run non-metric MDS on datasets with large sample size, we implemented a simple stochastic approximation of both metric and non-metric MDS (Listing \ref{lst:stochasticMDS}). We first describe the algorithmic steps used in both setups. In each training epoch, we randomly split the dataset into $\lfloor n/b \rfloor$ mini-batches of size $b$. If the input data are given as $\mathbf X$ matrix, then for each mini-batch, we compute $b\times b$ values of $D_{ij}$ (in order to avoid storing the entire $n\times n$ matrix $\mathbf D$ in memory, as this can be prohibitive for large $n$). Then we use the average stress (Eq.~\ref{eq:stress}) as the loss, with the sum only running over the $b^2$ terms in the mini-batch.

The $D_{ij}$ values in each mini-batch are divided by a normalization coefficient such that their mean squared off-diagonal value equals 1 (as in scikit-learn's non-metric MDS). This makes the optimization independent of the scale of the input data, so that the same learning rate and initialization scale can be used for all datasets. The coefficient is computed on the first mini-batch of each epoch and used for all mini-batches of that epoch, so that the target scale does not change from batch to batch. Note that the resulting scale discrepancy between our stochastic metric MDS and scikit-learn's metric MDS does not affect the stress-1 values we report, as they are computed after optimally rescaling the embedding (Section~\ref{sec:background}).

The entire model is an instance of \texttt{torch.nn.Embedding}, initialized with PCA (in case $\mathbf X$ is used as input) or with classical MDS (in case $\mathbf D$ is used). We scale the initialization to have a pre-specified standard deviation $\sigma_\text{init}$ of the first component. For optimization we use SGD with learning rate $1/(2b)$ (as in SMACOF) and cosine learning-rate decay over $n_\text{epochs}$. This is in agreement with \citet{agrafiotis2003stochastic} and \citet{zheng2018graph} who argued that starting with a large learning rate and decaying it down to zero helps optimizing stochastic metric MDS. The embedding is initialized with \texttt{sparse=True} and SGD does not use momentum so that each batch only updates its own coordinates.

For the stochastic version of non-metric MDS, the only modification to the above is that $D_{ij}$ values are transformed with isotonic regression (we use \texttt{IsotonicRegression} from scikit-learn) before they are normalized. As isotonic regression can be slow, we opted to fit it only on the first batch in each epoch; for subsequent batches in the same epoch, we use the fitted regression model to transform the $D_{ij}$ values. Fitting isotonic regression more often led to only very minor improvements in the final loss (Table~\ref{tab:isotonic-frequency}).

\paragraph{Default hyperparameters}

We use $\sigma_\text{init} = 0.5$ which we found to work better than $1$ on some of the datasets. For small datasets with $n\le5\,000$, we use batch size $b=100$ for non-metric MDS, and in other cases we use $b=1\,000$. The default number of epochs is set to $n_\text{epochs}=200$ if $n\le50\,000$ and to $n_\text{epochs}=20$ otherwise. We found these parameters to work well across all considered datasets. We increased the number of epochs to $n_\text{epochs}=500$ for non-metric MDS of the Tasic dataset, see below. Note that on some of our datasets a much smaller number of training epochs would suffice to reach a similar final loss.

\paragraph{Implementation details}

We observed that larger batch sizes were beneficial and allowed to achieve lower loss. However, the runtime of our implementation scales as $b^2 \cdot \lfloor n/b \rfloor = \mathcal O(nb)$, and in the non-metric case, the isotonic regression also becomes slower with larger $b$. We found $b=1000$ to be a reasonable compromise.


All experiments were run on a laptop with Intel Core i5-8265U CPU at 1.6~GHz and 16~Gb RAM.

Stress-1 was computed on all pairs for $n \le 5000$ and on a random subsample of 5000 points for larger datasets. We ran all experiments for three random seeds and report means and standard deviations.


\section{Experimental results}
\label{sec:experiments}

\subsection{Non-metric MDS mitigates the curse of dimensionality}

\begin{figure*}[t]
\includegraphics{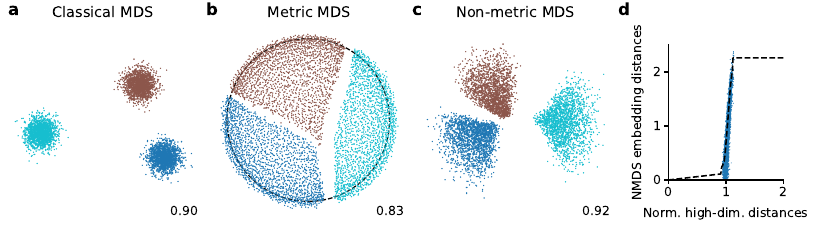}
\caption{Classical, metric, and non-metric MDS embeddings of the noisy blobs dataset ($n=5\,000$ in 1002 dimensions). Values in the lower-right are triplet accuracies (see text). \textbf{(b--c)} Embeddings obtained using the SMACOF implementation in scikit-learn with our improvements. The dashed circle in (b) shows our theoretical prediction and has radius $\bar D\pi/4$, where $\bar D$ is the average high-dimensional pairwise distance. \textbf{(d)} Shepard diagram for the non-metric MDS: embedding distances plotted against normalized high-dimensional distances. Dashed line shows monotonic regression fit.}
\label{fig:blobs}
\vspace{\floatsep}
\includegraphics{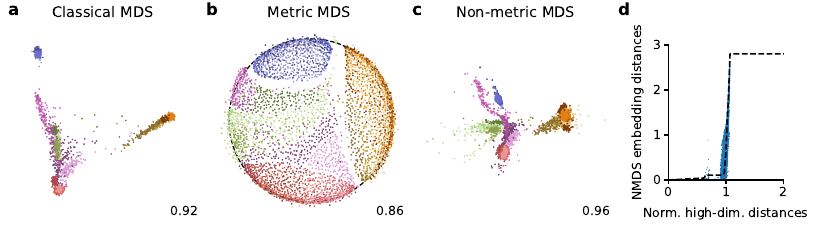}
\caption{Classical, metric, and non-metric MDS embeddings of the 1000 Genomes dataset ($n=3450$ in 53\,999 dimensions). See caption of Figure~\ref{fig:blobs} for explanations. Colors correspond to the sampling population (26 global populations), see \citet{1000genomes2015global}. The dashed circle in (b) shows our theoretical prediction and has radius $\bar D\pi/4$. }
\label{fig:genomes}
\end{figure*}

We started with analyzing a synthetic noisy blob dataset, with 5\,000 points sampled from a mixture of three well-separated spherical 2D Gaussians (with means sampled from $[-10, 10]$ and variances 1) with additional 1000 non-discriminative standard Gaussian dimensions (Figure~\ref{fig:blobs}). Classical MDS (i.e., PCA) produced an embedding with well-separated classes (Figure~\ref{fig:blobs}a), essentially recovering the original 2D data. At the same time, metric MDS produced only a weak class separation (Figure~\ref{fig:blobs}b) and in fact typically failed to yield separated classes when using random initialization (Figure~\ref{fig:random-init}; by default we used classical MDS as initialization for metric MDS). The stress-1 value 0.40 (Table~\ref{tab:stress}) was very close to the prediction of Thm.~\ref{thm:main} of $\sqrt{S^*}\approx 0.42$.

To evaluate the embeddings in a loss-independent way, we used triplet accuracy (for each sample A we randomly selected samples B and C from the upper/lower halves of pairwise distance distribution from A; we then computed the fraction of such triplets where A--C embedding distance was smaller than A--B). We also used Pearson and Spearman correlations between pairwise distances in the embedding space and in the original space (here we randomly sampled 5000 points and computed all distances within this subset). These are commonly used metrics to assess global embedding quality \citep[e.g.][]{becht2019dimensionality,kobak2019art,huang2022towards,bodt2025lowdim}.

On the noisy blob dataset, the triplet accuracy was higher for classical MDS (0.90) than for metric MDS (0.83). The metric MDS had the shape predicted above in Section~\ref{sec:curse}: it was mostly contained within the circle with radius $\bar D \pi/4$ (Figure~\ref{fig:blobs}b, dashed line), with higher density along the circle and lower density inside it. Indeed, the high-dimensional distances in this dataset were tightly concentrated around $\bar D\approx \sqrt{2\cdot 1000}$ (Figure~\ref{fig:blobs}d), making it impossible to reproduce them in 2D.

Using the scikit-learn implementation of non-metric MDS with our improvements (see Section~\ref{sec:background}), we found that non-metric MDS resulted in a clearer class separation (Figure~\ref{fig:blobs}c) and the highest triplet accuracy (0.92). Isotonic regression transformed tightly concentrated high-dimensional distances to a range between 0 and 2 (Figure~\ref{fig:blobs}d; in the MDS literature this kind of plot is known as \textit{Shepard diagram}). Pearson and Spearman evaluations yielded the same conclusions (Tables~\ref{tab:pearson}--\ref{tab:spearman}).

We observed the same behavior in the 1000 Genomes dataset \citep{1000genomes2015global}, where we used the preprocessed data from \citet{bodt2025lowdim}, resulting in 53\,999 feature dimensions which we converted to the distance matrix (Figure~\ref{fig:genomes}). Here input distances were also highly concentrated (Figure~\ref{fig:genomes}d). As a result, metric MDS struggled to represent the data and produced a nearly spherical embedding with triplet accuracy (0.86) lower than that of classical MDS (0.92). The radius of the metric MDS embedding was in good agreement with our theoretical prediction (Figure~\ref{fig:genomes}b) as was the stress value of $0.41\approx \sqrt{S^*}$ (Table~\ref{tab:stress}). In contrast, non-metric MDS produced a more informative embedding with better class separation and much higher triplet accuracy (0.96). In genomics it is common to pre-process the data with PCA. Doing it prior to metric MDS improved the triplet accuracy of the resulting embedding (50 PCA dimensions: 0.89; 10 dimensions: 0.91) but did not reach the level of non-metric MDS. 

On the other hand, on the Digits dataset (available in scikit-learn) there was little difference between the performance of metric and non-metric MDS (Figure~\ref{fig:digits}). This makes sense because the pairwise distances in this dataset were less tightly concentrated (Figure~\ref{fig:digits}d).

\subsection{Stochastic non-metric MDS is accurate and fast}

Using all three datasets presented above (noisy blobs, 1000 Genomes, and Digits), we compared our stochastic implementation with the scikit-learn implementation with our modifications. We found that both implementations achieved the same stress-1 values on all three datasets and for both metric/non-metric MDS (Table~\ref{tab:stress}, left). At the same time, the stochastic implementation was much faster in all cases (Table~\ref{tab:stress}, right). The speed-up was particularly noticeable for non-metric MDS.

\begin{table}[t]
\caption{Stress-1 values and runtimes using the SMACOF and SGD implementations. On all datasets, our stochastic implementation achieves the same embedding quality as the reference non-stochastic implementation while being much faster. Tasic and MNIST datasets are too large for the non-stochastic implementation. The runtimes correspond to the setup described in the text. All stress-1 values for SGD have stds 0.00; all runtimes for SGD have stds at most 1s (1m).}
\label{tab:stress}
\begin{center}
\setlength{\tabcolsep}{4pt}
\begin{tabular}{lccccc|ccccc}
\toprule
& \multicolumn{5}{c|}{\textbf{Stress-1}} & \multicolumn{5}{c}{\textbf{Runtime}} \vspace{.5em}\\
& Blobs & Gen. & Digits & Tasic & MNIST & Blobs & Gen. & Digits & Tasic & MNIST \\ 
$n$ &  &  &  &  &  & 5\,000 & 3\,450 & 1\,797 & 23\,822 & 70\,000 \\
\midrule
Metric & 0.40 & 0.41 & 0.33 & -- & -- & 359 s & 358 s & 15 s & -- & -- \\
\hspace{1em}with SGD & 0.40 & 0.41 & 0.33 & 0.25 & 0.36 & 21 s & 19 s & 5 s & 1 m & 2 m \\
\midrule
Non-metric & 0.24 & 0.13 & 0.28 & -- & -- & 1086 s & 473 s & 152 s & -- & -- \\
\hspace{1em}with SGD & 0.24 & 0.13 & 0.28 & 0.06 & 0.30 & 18 s & 23 s & 8 s & 23 m & 3 m \\
\bottomrule
\end{tabular}
\end{center}

\vspace{\floatsep}

\caption{Triplet accuracies. The highest value in each column is highlighted in bold, together with all values within 0.02. Non-metric MDS always achieves the best (or one of the best) result(s). See Tables~\ref{tab:pearson}--\ref{tab:spearman} for Pearson and Spearman correlations between pairwise distances. Scikit-learn implementation for the first three datasets, our stochastic SGD for the last two.}
\label{tab:triplet}
\begin{center}
\begin{tabular}{lccccc}
\toprule
& Blobs & Gen. & Digits & Tasic & MNIST \\
\midrule
Classical & 0.90 {\scriptsize$\pm$.00} & 0.92 {\scriptsize$\pm$.00} & 0.80 {\scriptsize$\pm$.00} & \textbf{0.99} {\scriptsize$\pm$.00} & 0.72 {\scriptsize$\pm$.00} \\
Metric & 0.83 {\scriptsize$\pm$.00} & 0.86 {\scriptsize$\pm$.00} & \textbf{0.84} {\scriptsize$\pm$.00} & 0.94 {\scriptsize$\pm$.00} & 0.77 {\scriptsize$\pm$.00} \\
Non-metric & \textbf{0.92} {\scriptsize$\pm$.00} & \textbf{0.96} {\scriptsize$\pm$.00} & \textbf{0.85} {\scriptsize$\pm$.00} & \textbf{0.99} {\scriptsize$\pm$.00} & \textbf{0.82} {\scriptsize$\pm$.00} \\
\midrule
$t$-SNE & 0.81 {\scriptsize$\pm$.04} & 0.79 {\scriptsize$\pm$.01} & 0.75 {\scriptsize$\pm$.00} & 0.79 {\scriptsize$\pm$.00} & 0.65 {\scriptsize$\pm$.01} \\
UMAP & 0.85 {\scriptsize$\pm$.02} & 0.82 {\scriptsize$\pm$.03} & 0.67 {\scriptsize$\pm$.01} & 0.82 {\scriptsize$\pm$.03} & 0.64 {\scriptsize$\pm$.00} \\
PaCMAP & 0.86 {\scriptsize$\pm$.00} & 0.80 {\scriptsize$\pm$.01} & 0.74 {\scriptsize$\pm$.00} & 0.86 {\scriptsize$\pm$.01} & 0.64 {\scriptsize$\pm$.00} \\
\bottomrule
\end{tabular}
\end{center}
\end{table}

The scikit-learn implementation constructs the full $\mathbf D$ matrix in memory, which becomes prohibitive for $n \gg 10\,000$. Due to memory overflow, we were unable to run it on our two further datasets: the Tasic dataset of single-cell transcriptomic data from mouse cortex \citep{tasic2018shared}, analyzed previously in \citet{kobak2019art} and \citet{bodt2025lowdim} (we followed the preprocessing from these two papers), and the MNIST dataset. Therefore, for these two datasets, all following results were obtained using our stochastic implementation.

\begin{figure*}[t]
\includegraphics{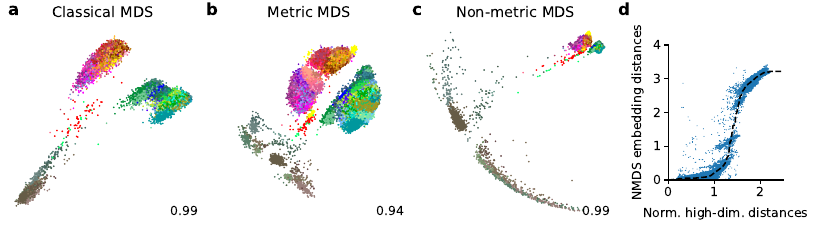}
\caption{Classical, metric, and non-metric MDS embeddings of the Tasic dataset ($n=23\,822$ in 50 dimensions). See caption of Figure~\ref{fig:blobs} for explanations. Colors correspond to cell types (warm colors: inhibitory neurons; cold colors: excitatory neurons; brown/grey colors: non-neural cells). See Figure~\ref{fig:tasic-shepard} for the same Shepard diagram as in (d), but colored by cell identity of cell pairs.}
\label{fig:tasic}
\end{figure*}

On the Tasic dataset, non-metric MDS produced an embedding with the shared highest triplet accuracy (0.99, as for PCA, vs. 0.94 for metric MDS) and the highest Spearman correlation (0.95, vs. 0.91--0.92 for PCA and metric MDS) (Figure~\ref{fig:tasic}). One salient feature of this dataset is that it contains a small amount of various non-neural cells that transcriptomically are (a) far away from all neural cells; and (b) far away from each other. Non-metric MDS emphasized this structure by locating non-neural cells along an arc going around the cluster of neural cells. This happened at a price of squeezing all neural cells together, compared to the metric MDS embedding. Note that here we increased the $n_\text{epochs}$ value for non-metric MDS to 500; our default value of 200 produced a similar-looking embedding but with a less pronounced arc structure.

On the MNIST dataset, the difference between metric and non-metric MDS was visually not very apparent (Figure~\ref{fig:mnist}), but non-metric MDS did show higher global structure preservation, especially for the distance correlations (Tables~\ref{tab:pearson}--\ref{tab:spearman}). Similar to the Digits dataset, the MNIST dataset arguably lacks global structure that can be represented by any MDS. One aspect of the data visible in the non-metric MDS embedding is the higher density of digits `1' (orange color in Figure~\ref{fig:mnist}c).

\subsection{Across datasets, non-metric MDS achieves the highest global structure preservation}

Across all five datasets, non-metric MDS achieved the highest performance in 14 out of 15 cases (five datasets and three evaluation metrics), including a shared highest in 3 cases (Tables~\ref{tab:triplet}, \ref{tab:pearson}, \ref{tab:spearman}).

For completeness, we also evaluated the $t$-SNE and the UMAP embeddings of all five datasets. These two algorithms are currently by far the most popular methods of representing high-dimensional data in 2D \citep{bodt2025lowdim}. Both $t$-SNE and UMAP are designed to preserve the nearest neighbors, which makes them excel at local structure preservation. Indeed, these two algorithms strongly outperformed all flavors of MDS in terms of \textit{local} quality metrics, such as $k$NN recall and $k$NN accuracy (Tables~\ref{tab:knn_recall}--\ref{tab:knn_acc}). $t$-SNE and UMAP were also much faster than our stochastic non-metric MDS implementation (Table~\ref{tab:runtimes}). At the same time, their capacity to preserve \textit{global} structure of the data is known to be very poor \citep{bodt2025lowdim} and indeed, this is what we confirmed on our datasets (Tables~\ref{tab:triplet}, \ref{tab:pearson}, \ref{tab:spearman}). We have also evaluated PaCMAP \citep{wang2021understanding} as the neighbor-embedding method designed to preserve global structure as well; however, its triplet performance was not consistently better than that of $t$-SNE and UMAP (Table~\ref{tab:triplet}).

\section{Discussion}

Our contribution in this paper is two-fold. Conceptually, we showed that metric MDS suffers from the curse of dimensionality. We derived the a perturbation bound stating that the global optimum of metric MDS remains close to a circular embedding under high-dimensional noise. We confirmed this effect in simulations and real datasets using \texttt{scikit-learn}'s implementation of metric MDS.

Practically, we developed a fast implementation of non-metric MDS, showed that non-metric MDS is more robust than metric MDS and can strongly outperform other embedding methods in terms of global structure preservation, in particular on datasets with prominent high-dimensional noise. Non-metric MDS was introduced over 60 years ago but has received little attention in modern manifold-learning and dimensionality-reduction literature. We are not aware of any modern benchmark studies of dimensionality-reduction methods that include non-metric MDS or of any prior scalable implementations. 

\paragraph{Future work and possible applications} 

Several recent studies have developed hybrid methods combining neighbor-embedding methods such as $t$-SNE \citep{maaten2008visualizing} or UMAP \citep{mcinnes2018umap} that exhibit high local structure preservation with global methods such as PCA or MDS. For example, SQuadMDS-hybrid \citep{lambert2022squadmds} includes both $t$-SNE and metric MDS losses in its composite loss term. Similarly, DREAMS \citep{kury2025dreams} regularized $t$-SNE towards a pre-computed PCA or MDS embedding. Our results suggest that non-metric MDS can be a good choice for a global counterpart in such hybrid methods. 
Other examples of modern methods relying on metric MDS are PHATE~\citep{moon2019visualizing} and HeatGeo~\citep{huguet2023heat} which approximate diffusion-based graph distances in 2D.
Here also, non-metric MDS could provide additional flexibility to embed the diffusion-based distances.

A recent work by \citet{li2025euclidean} suggested a deflation procedure that can remove the contribution of high-dimensional noise from a distance matrix. It would be interesting to study if this can be combined with metric MDS to achieve similar results as with non-metric MDS.

\clearpage

\subsection*{AI use statement}

In this work, we used generative AI tools (Claude and ChatGPT) to 
formulate mathematical claims, to provide critical ingredients for the proofs, and to write down the final proofs. We used interactively sessions with Claude 4.7 for Theorem~\ref{thm:main} and Appendix~\ref{sec:proofs} and ChatGPT 5.5 for Proposition~\ref{prop:perturbation_bound} and Appendix~\ref{app:perturbation}. We manually and in detail checked the correctness of all generated proofs. For instance, Claude's initial proof of Theorem~\ref{thm:main} only showed that the KKT conditions lead to the Riesz equilibrium measure, but did not verify that it actually satisfies the KKT conditions. Moreover, there was initially no convexity argument and Claude only claimed that $\rho^*$ is optimal among radially symmetric distributions. When prompted to verify the off-support KKT condition, Claude hallucinated a \textit{minimum} principle for subharmonic functions. But when we pushed back, it came up with the correct solution based on the \textit{maximum} principle for subharmonic functions, which indeed exists and yields a valid proof.

All code for the initial submission was implemented manually by the authors. During revision, we used Claude Sonnet 5.5 and Opus 5.5 to refactor the codebase and to extend our benchmark with additional evaluations and methods.

The manuscript text was written by hand.



\subsection*{Data and code}

All our code is available at \url{https://github.com/kobaklab/nmds}. All used datasets are openly available online, with links in the Github README files. 







\bibliography{mds}
\bibliographystyle{iclr2027_conference}

\clearpage

\appendix

\renewcommand{\thefigure}{S\arabic{figure}}
\renewcommand{\theHfigure}{S\arabic{figure}}
\setcounter{figure}{0}  
\renewcommand{\thetable}{S\arabic{table}}
\renewcommand{\theHtable}{S\arabic{table}}
\setcounter{table}{0} 
\renewcommand{\thelstlisting}{S\arabic{lstlisting}}
\renewcommand{\theHlstlisting}{S\arabic{lstlisting}}
\setcounter{lstlisting}{0} 

\section{Supplementary tables}

\begin{table}[h!]
\caption{Runtime and stress-1 of our stochastic non-metric MDS implementation depending on how often the isotonic regression is run. Results are shown for the Blobs and the MNIST datasets. All hyperparameters (batch size, number of epochs, etc.) are kept constant at the values described in the main text (Blobs: batch size 100, sample size 5000, 200 epochs; MNIST: batch size 1000, sample size 70\,000, 20 epochs).}
\label{tab:isotonic-frequency}
\begin{center}
\begin{tabular}{lcc|cc}
\toprule
& \multicolumn{2}{c|}{Blobs} & \multicolumn{2}{c}{MNIST} \\ 
& Time & Stress-1 & Time & Stress-1 \\
\midrule
Once every 100 epochs & 22 {\scriptsize$\pm$0} s & 0.245 {\scriptsize$\pm$.000} & --- & --- \\
Once every 10 epochs & 22 {\scriptsize$\pm$0} s & 0.244 {\scriptsize$\pm$.000} & 182 {\scriptsize$\pm$2} s & 0.303 {\scriptsize$\pm$.002} \\
Once every epoch (default) & 23 {\scriptsize$\pm$1} s & 0.244 {\scriptsize$\pm$.000} & 192 {\scriptsize$\pm$2} s & 0.298 {\scriptsize$\pm$.002} \\
Once every 10 batches & 26 {\scriptsize$\pm$1} s & 0.244 {\scriptsize$\pm$.000} & 235 {\scriptsize$\pm$0} s & 0.299 {\scriptsize$\pm$.002} \\
Every batch & 55 {\scriptsize$\pm$0} s & 0.244 {\scriptsize$\pm$.000} & 654 {\scriptsize$\pm$2} s & 0.299 {\scriptsize$\pm$.002} \\
\bottomrule
\end{tabular}
\end{center}
\end{table}

\begin{table}[h!]
\caption{Pearson correlation coefficient between pairwise distances in the high-dimensional space and in the embedding. The highest value in each column is highlighted in bold, together with all values within 0.02. Deterministic runs are shown without stds. Non-metric MDS always achieves the best (or one of the best) result(s).}
\label{tab:pearson}
\begin{center}
\begin{tabular}{lccccc}
\toprule
& Blobs & Gen. & Digits & Tasic & MNIST \\
\midrule
Classical & 0.76 & 0.83 & 0.59 & \textbf{0.92} {\scriptsize$\pm$.00} & 0.52 {\scriptsize$\pm$.01} \\
Metric & 0.67 & 0.70 & \textbf{0.73} & 0.90 {\scriptsize$\pm$.00} & 0.62 {\scriptsize$\pm$.00} \\
Non-metric & \textbf{0.85} & \textbf{0.91} & \textbf{0.74} & 0.87 {\scriptsize$\pm$.00} & \textbf{0.76} {\scriptsize$\pm$.00} \\
\midrule
$t$-SNE & 0.68 {\scriptsize$\pm$.04} & 0.66 {\scriptsize$\pm$.01} & 0.56 {\scriptsize$\pm$.00} & 0.46 {\scriptsize$\pm$.01} & 0.36 {\scriptsize$\pm$.01} \\
UMAP & 0.69 {\scriptsize$\pm$.05} & 0.71 {\scriptsize$\pm$.05} & 0.43 {\scriptsize$\pm$.02} & 0.50 {\scriptsize$\pm$.05} & 0.34 {\scriptsize$\pm$.01} \\
PaCMAP & 0.68 {\scriptsize$\pm$.00} & 0.64 {\scriptsize$\pm$.01} & 0.54 {\scriptsize$\pm$.00} & 0.56 {\scriptsize$\pm$.01} & 0.34 {\scriptsize$\pm$.00} \\
\bottomrule
\end{tabular}
\end{center}
\end{table}

\begin{table}[h!]
\caption{Spearman correlation coefficient between pairwise distances in the high-dimensional space and in the embedding. The highest value in each column is highlighted in bold, together with all values within 0.02. Deterministic runs are shown without stds. Non-metric MDS always achieves the best (or one of the best) result(s).}
\label{tab:spearman}
\begin{center}
\begin{tabular}{lccccc}
\toprule
& Blobs & Gen. & Digits & Tasic & MNIST \\
\midrule
Classical & 0.78 & 0.84 & 0.58 & 0.91 {\scriptsize$\pm$.00} & 0.49 {\scriptsize$\pm$.01} \\
Metric & 0.68 & 0.71 & 0.73 & 0.91 {\scriptsize$\pm$.00} & 0.62 {\scriptsize$\pm$.00} \\
Non-metric & \textbf{0.85} & \textbf{0.96} & \textbf{0.76} & \textbf{0.95} {\scriptsize$\pm$.00} & \textbf{0.78} {\scriptsize$\pm$.00} \\
\midrule
$t$-SNE & 0.61 {\scriptsize$\pm$.09} & 0.62 {\scriptsize$\pm$.01} & 0.51 {\scriptsize$\pm$.00} & 0.55 {\scriptsize$\pm$.00} & 0.33 {\scriptsize$\pm$.01} \\
UMAP & 0.70 {\scriptsize$\pm$.04} & 0.62 {\scriptsize$\pm$.07} & 0.37 {\scriptsize$\pm$.02} & 0.58 {\scriptsize$\pm$.02} & 0.30 {\scriptsize$\pm$.01} \\
PaCMAP & 0.69 {\scriptsize$\pm$.00} & 0.62 {\scriptsize$\pm$.01} & 0.49 {\scriptsize$\pm$.00} & 0.67 {\scriptsize$\pm$.01} & 0.30 {\scriptsize$\pm$.00} \\
\bottomrule
\end{tabular}
\end{center}
\end{table}

\begin{table}[h!]
\caption{$k$NN recall, that is the fraction of $k=10$ original nearest neighbors that are among the $k=10$ nearest neighbors in the embedding (evaluated on a subset of 5000 query points for datasets with $n>5000$). The highest value in each column is highlighted in bold, together with all values within 0.02. Deterministic runs are shown without stds. As expected, $t$-SNE always achieves the best result (on Blobs, within-class noise is too large and high-dimensional neighbors are near-random).}
\label{tab:knn_recall}
\begin{center}
\begin{tabular}{lccccc}
\toprule
& Blobs & Gen. & Digits & Tasic & MNIST \\
\midrule
Classical & \textbf{0.01} & 0.04 & 0.12 & 0.02 {\scriptsize$\pm$.00} & 0.01 {\scriptsize$\pm$.00} \\
Metric & \textbf{0.01} & 0.04 & 0.15 & 0.04 {\scriptsize$\pm$.00} & 0.01 {\scriptsize$\pm$.00} \\
Non-metric & \textbf{0.01} & 0.04 & 0.13 & 0.03 {\scriptsize$\pm$.00} & 0.01 {\scriptsize$\pm$.00} \\
\midrule
$t$-SNE & \textbf{0.02} {\scriptsize$\pm$.00} & \textbf{0.20} {\scriptsize$\pm$.00} & \textbf{0.59} {\scriptsize$\pm$.00} & \textbf{0.45} {\scriptsize$\pm$.00} & \textbf{0.36} {\scriptsize$\pm$.00} \\
UMAP & \textbf{0.03} {\scriptsize$\pm$.00} & 0.14 {\scriptsize$\pm$.00} & 0.49 {\scriptsize$\pm$.00} & 0.25 {\scriptsize$\pm$.00} & 0.10 {\scriptsize$\pm$.00} \\
PaCMAP & \textbf{0.01} {\scriptsize$\pm$.00} & 0.11 {\scriptsize$\pm$.00} & 0.44 {\scriptsize$\pm$.01} & 0.21 {\scriptsize$\pm$.00} & 0.08 {\scriptsize$\pm$.00} \\
\bottomrule
\end{tabular}
\end{center}
\end{table}

\begin{table}[h!]
\caption{$k$NN accuracy ($k=10$) in the embedding. The highest value in each column is highlighted in bold, together with all values within 0.02. Deterministic runs are shown without stds. As expected, $t$-SNE always achieves the best or one of the best result(s).}
\label{tab:knn_acc}
\begin{center}
\begin{tabular}{lccccc}
\toprule
& Blobs & Gen. & Digits & Tasic & MNIST \\
\midrule
Classical & \textbf{1.00} & 0.56 {\scriptsize$\pm$.02} & 0.65 {\scriptsize$\pm$.02} & 0.21 {\scriptsize$\pm$.00} & 0.43 {\scriptsize$\pm$.01} \\
Metric & \textbf{1.00} & 0.49 {\scriptsize$\pm$.01} & 0.72 {\scriptsize$\pm$.01} & 0.44 {\scriptsize$\pm$.01} & 0.47 {\scriptsize$\pm$.00} \\
Non-metric & \textbf{1.00} & 0.59 {\scriptsize$\pm$.01} & 0.69 {\scriptsize$\pm$.02} & 0.23 {\scriptsize$\pm$.01} & 0.48 {\scriptsize$\pm$.00} \\
\midrule
$t$-SNE & \textbf{1.00} {\scriptsize$\pm$.00} & \textbf{0.89} {\scriptsize$\pm$.01} & \textbf{0.99} {\scriptsize$\pm$.03} & \textbf{0.90} {\scriptsize$\pm$.01} & \textbf{0.97} {\scriptsize$\pm$.00} \\
UMAP & \textbf{1.00} {\scriptsize$\pm$.00} & 0.83 {\scriptsize$\pm$.01} & \textbf{0.99} {\scriptsize$\pm$.02} & \textbf{0.90} {\scriptsize$\pm$.00} & \textbf{0.97} {\scriptsize$\pm$.00} \\
PaCMAP & \textbf{1.00} {\scriptsize$\pm$.00} & 0.85 {\scriptsize$\pm$.01} & \textbf{0.99} {\scriptsize$\pm$.02} & \textbf{0.89} {\scriptsize$\pm$.00} & \textbf{0.98} {\scriptsize$\pm$.00} \\
\bottomrule
\end{tabular}
\end{center}
\end{table}

\begin{table}[h!]
\caption{Runtimes of all methods. Deterministic runs are shown without stds. openTSNE and UMAP are run single-threaded.}
\label{tab:runtimes}
\begin{center}
\begin{tabular}{lrrrrr}
\toprule
& Blobs & Gen. & Digits & Tasic & MNIST \\
\midrule
Classical & 2 s & 23 s & 0 s & 0 m & 0 m \\
Metric (SMACOF) & 359 s & 358 s & 15 s & -- & -- \\
Metric (SGD) & 21 {\scriptsize$\pm$0} s & 19 {\scriptsize$\pm$0} s & 5 {\scriptsize$\pm$0} s & 1 {\scriptsize$\pm$0} m & 1 {\scriptsize$\pm$0} m \\
Non-metric (SMACOF) & 1086 s & 473 s & 152 s & -- & -- \\
Non-metric (SGD) & 18 {\scriptsize$\pm$0} s & 23 {\scriptsize$\pm$0} s & 8 {\scriptsize$\pm$0} s & 23 {\scriptsize$\pm$1} m & 3 {\scriptsize$\pm$0} m \\
\midrule
$t$-SNE & 35 {\scriptsize$\pm$0} s & 23 {\scriptsize$\pm$0} s & 7 {\scriptsize$\pm$0} s & 1 {\scriptsize$\pm$0} m & 5 {\scriptsize$\pm$0} m \\
UMAP & 7 {\scriptsize$\pm$0} s & 9 {\scriptsize$\pm$1} s & 4 {\scriptsize$\pm$0} s & 0 {\scriptsize$\pm$0} m & 2 {\scriptsize$\pm$0} m \\
PaCMAP & 6 {\scriptsize$\pm$0} s & 45 {\scriptsize$\pm$1} s & 1 {\scriptsize$\pm$0} s & 0 {\scriptsize$\pm$0} m & 1 {\scriptsize$\pm$0} m \\
\bottomrule
\end{tabular}
\end{center}
\end{table}

\clearpage

\subsection{Supplementary figures}    

\begin{figure}[h]
\begin{center}
\includegraphics{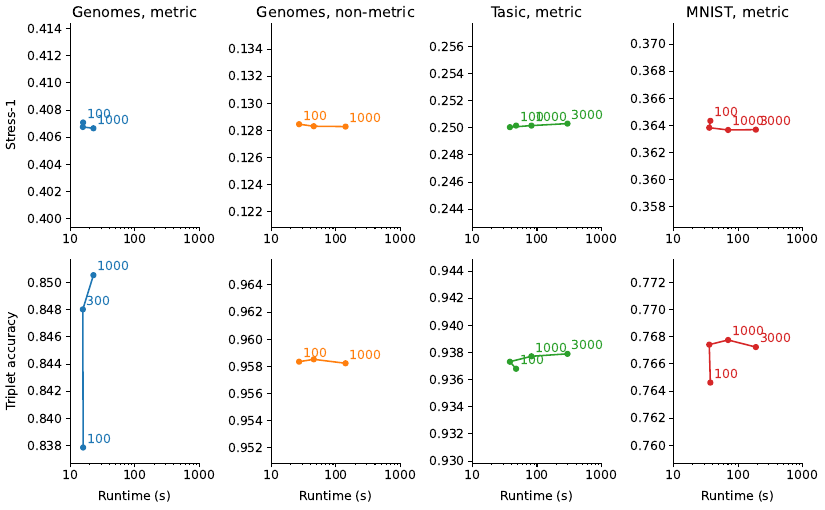}
\end{center}
\caption{Stress-1 and triplet accuracies obtained using the SGD implementation of metric MDS using different batch sizes. All y-axis ranges are fixed to 0.015.}
\label{fig:sgd-params}
\end{figure}


\begin{figure}[h]
  \centering
  \includegraphics[width=\linewidth]{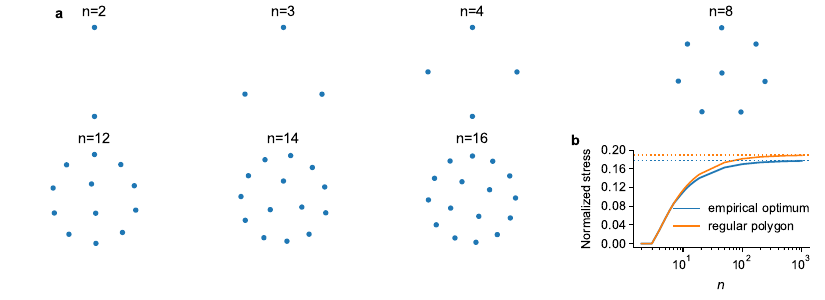}
  \caption{Empirically optimal 2D MDS layouts of $n$ equidistant points, obtained using the scikit-learn implementation. \textbf{(a)} For $n=2$ and $3$ exact equidistance can be achieved. Starting with $n=4$ pairwise equidistance needs to be approximated. Empirically, for $n=4,\dots, 7$ a regular polygon is optimal. At $n\ge8$ our experiments showed that adding additional points inside the polygon achieves lower loss. \textbf{(b)} Average stress values for the regular polygons and MDS embeddings (averaged over 3 seeds, standard deviations were negligible), as a function of sample size $n$. Dotted lines show the optimal stress for the continuum limits from Lemma~\ref{lem:opt_circle} and Theorem~\ref{thm:main}.}
  \label{fig:reg_poly}
\end{figure}

\begin{figure}[ht]
\includegraphics[width=\linewidth]{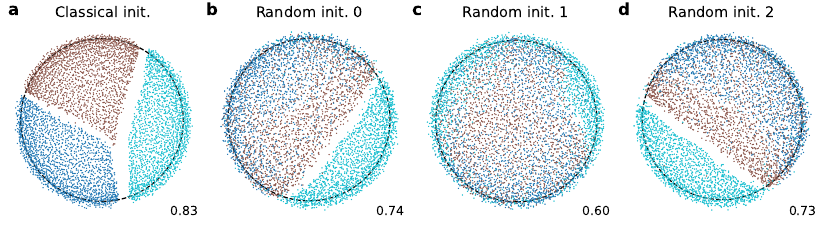}
\caption{Metric MDS embedding of the noisy blobs dataset with the classical MDS and random initializations.}
\label{fig:random-init}
\end{figure}

\begin{figure}[ht]
\includegraphics[width=\linewidth]{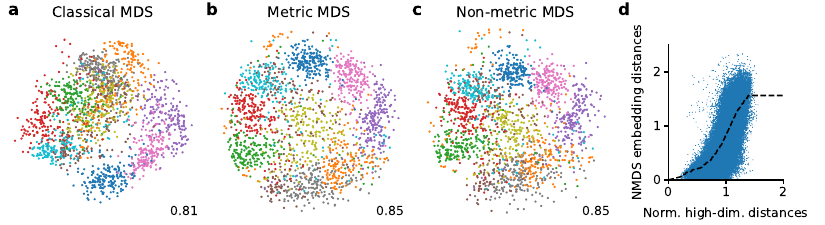}
\caption{Classical, metric, and non-metric MDS embeddings of the Digits dataset ($n=1797$ in 64 dimensions). See caption of Figure~\ref{fig:blobs} for explanations.}
\label{fig:digits}
\end{figure}

\begin{figure*}[ht]
\includegraphics{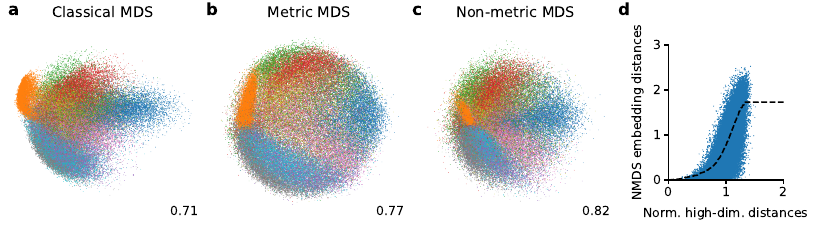}
\caption{Classical, metric, and non-metric MDS embeddings of the MNIST dataset ($n=70\,000$ in 784 dimensions). See caption of Figure~\ref{fig:blobs} for explanations.}
\label{fig:mnist}
\end{figure*}

\begin{figure}[ht]
\begin{center}
\includegraphics{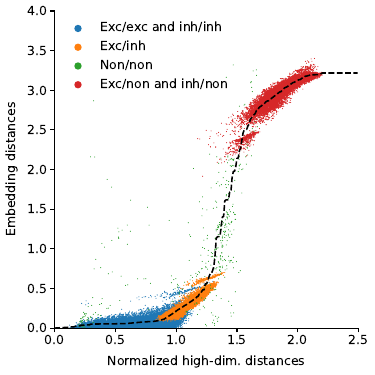}
\end{center}
\caption{Shepard diagram for the non-metric MDS embedding of the Tasic dataset. Reproduced from Figure~\ref{fig:tasic}d but here each pair of cells is colored based on the cells' identities: excitatory/excitatory and inhibitory/inhibitory neuron pairs are shown in blue; excitatory/inhibitory neuron pairs --- in orange; non-neural/non-neural pairs --- in  green; neural/non-neural pairs --- in red.}
\label{fig:tasic-shepard}
\end{figure}

\vfill


\clearpage

\subsection{Supplementary listing}    

\begin{lstlisting}[language=Python, caption=Stochastic MDS implementation.,label={lst:stochasticMDS}]
import numpy as np
import time
import torch
from sklearn.manifold import ClassicalMDS
from sklearn.decomposition import PCA
from sklearn.isotonic import IsotonicRegression
from scipy.spatial.distance import cdist

def stochasticMDS(
    X,
    n_components=2,
    precomputed_distances=False,
    non_metric=False,
    n_epochs="auto",
    batch_size="auto",
    print_every_epoch="auto",
    init_scale=0.5,
    init=None,
    seed=42,
    compute_all_distances="auto",
):

    ### Default hyperparameter values
    sample_size = X.shape[0]

    if batch_size == "auto":
        if sample_size <= 5000 and non_metric:
            batch_size = 100
        else:
            batch_size = 1000
        batch_size = min(batch_size, sample_size)

    if n_epochs == "auto":
        if sample_size < 50_000:
            n_epochs = 200
        else:
            n_epochs = 20

    if print_every_epoch == "auto":
        if n_epochs > 20:
            print_every_epoch = 10
        else:
            print_every_epoch = 1

    ### Initialization
    if init is not None:
        Z = init
    else:
        if precomputed_distances:
            Z = ClassicalMDS(
                n_components=n_components, metric="precomputed"
            ).fit_transform(X)
        else:
            Z = PCA(
                n_components=n_components, random_state=seed
            ).fit_transform(X)
        Z /= np.std(Z[:, 0])
        Z *= init_scale

    if compute_all_distances == "auto":
        compute_all_distances = sample_size <= 10_000
    if compute_all_distances and not precomputed_distances:
        X = euclidean_distances(X)
        precomputed_distances = True
    
    generator = torch.Generator().manual_seed(seed)

    ### Set up the model
    model = torch.nn.Embedding.from_pretrained(
        torch.tensor(np.ascontiguousarray(Z, dtype=np.float32)), freeze=False, sparse=True,
    )

    optimizer = torch.optim.SGD(
        model.parameters(), lr=1 / (2 * batch_size)
    )
    scheduler = torch.optim.lr_scheduler.CosineAnnealingLR(
        optimizer, T_max=n_epochs
    )

    if non_metric:
        ir = IsotonicRegression(out_of_bounds="clip")

    ### Training loop
    n_batches = sample_size // batch_size
    for epoch in range(n_epochs):
        shuffle = torch.randperm(sample_size, generator=generator)

        epoch_loss = 0
        for batch in range(n_batches):
            ind = shuffle[
                batch * batch_size :
                (batch + 1) * batch_size
            ]
            ind_np = ind.numpy()

            if precomputed_distances:
                dists = X[np.ix_(ind_np, ind_np)]
            else:
                batch_X = torch.from_numpy(X[ind_np])
+               dists = torch.cdist(batch_X, batch_X).fill_diagonal_(0).numpy()

            optimizer.zero_grad()
            embedding = model(ind)
            model_dists = torch.cdist(
                    embedding, embedding
            )

            ### Non-metric part
            if non_metric:
                if epoch >= start_nonmetric_from_epoch:
                    if batch == 0:
                        ir.fit(
                            dists.ravel(),
                            (
                                model_dists.detach().
                                numpy().ravel()
                            )
                        )

                    dists = ir.transform(
                        dists.ravel()
                    ).reshape(
                        dists.shape
                    )

            ### Normalization
            if batch == 0:
                dists_scaling = np.sqrt(
                    np.sum(dists ** 2) / (batch_size ** 2 - batch_size)
                )
            dists = dists / dists_scaling

            dists = torch.tensor(
                dists.astype(np.float32)
            )

            ### Loss function = average stress
            loss = 0.5 * torch.sum(
                (model_dists - dists) ** 2
            )

            loss.backward()
            optimizer.step()
            epoch_loss += loss.item()

        scheduler.step()
        epoch_loss /= n_batches

    Z = model.weight.detach().numpy()
    return Z, epoch_loss
\end{lstlisting}

\clearpage

\section{Extended robustness analysis}
\label{sec:extended_robustness}

In addition to the setup in Fig.~\ref{fig:robustness}, we also ran a larger suite of robustness experiments. In particular, we 
\begin{itemize}
    \item explored two dataset sizes, $n^*=99$ and $n^*=297$;
    \item made the total number of points random by sampling $n\sim \text{Poisson}(n^*)$;
    \item made the sizes of the three clusters random, by sampling their sizes from $\text{Multinomial}(n; 1/3, 1/3, 1/3)$;
    \item varied the pairwise distances between clusters in 10 geometric steps between $D=8$ and $D=64$;
    \item broke the noise-free equidistance of the clusters by jittering the points away from their cluster centers by standard Gaussians in the first two dimensions, scaled by $D/10$;
    \item used different standard deviations per noise dimension of $\sigma=1$ and $\sigma=2$;
    \item varied the type of noise in the additional $d$ noise dimension between Gaussian, uniform, and exponential (all with standard deviation $\sigma$);
    \item increased the range of noise dimensions to $d\in[20, 15\,000\,000]$.
\end{itemize}
All configurations are summarized in Tab.~\ref{tab:robustness_experiments}. We ran each $(d, D)$ configuration 10 times, each time with a different noise realization (and, where random, different cluster sizes and jitter) and a different random initialization of MDS, and averaged the $k$NN accuracy over the 10 runs. We then fitted a power law to the 10 points $\big(D, d^*(D/\sigma)\big)$, where $d^*$ is the smallest number of noise dimensions at which the averaged kNN accuracy dipped below $0.7$. For metric MDS, the fitted slopes were within [2.0, 2.3] and for non-metric MDS within [3.9, 4.1] in close alignment with our theoretical predictions of $2$ and $4$, respectively. 

\begin{table}[h]
\centering
\caption{Extended robustness analysis settings. The first line summarizes the setup for Fig.~\ref{fig:robustness}b, the other lines show the settings for the extended analysis. The last three columns are slopes for metric, non-metric, and classical MDS.}
\label{tab:robustness_experiments}
\begin{tabular}{lllllccc}
\toprule
Noise & $\sigma$ & Jitter & Size & Sampling & mMDS & nMDS & cMDS \\
\midrule
Gaussian & 1.0 & No & $=99$ & Fixed & 2.24 & 4.11 & 4.75 \\  
Gaussian & 1.0 & Yes & $\sim99$ & Random & 2.22 & 4.05 & 4.04 \\  
Uniform & 1.0 & Yes & $\sim99$ & Random & 2.22 & 3.98 & 4.07 \\  
Exponential & 1.0 & Yes & $\sim99$ & Random & 2.33 & 4.04 & 4.04 \\  
Gaussian & 2.0 & Yes & $\sim99$ & Random & 2.33 & 3.94 & 4.23 \\  
Gaussian & 1.0 & Yes & $\sim297$ & Random & 2.02 & 3.92 & 3.94 \\  
\bottomrule
\end{tabular}
\end{table}

\section{Optima of metric MDS for equidistant inputs}
\label{sec:proofs}

\begin{lemma}[Optimal regular polygons]
  \label{lem:finiteN}
  For $n \geq 2$ points placed on a regular $n$-gon of circumradius $r$,
    the stress-minimizing radius is
    \begin{equation*}
      r_n^* = \frac{1}{n}\cot\frac{\pi}{2n} \xrightarrow{n\to\infty} 2/\pi.
    \end{equation*}
    The corresponding optimal average stress is
    $$
    S^*_n = 1-\frac{2}{n(n-1)}\cot^2\left(\frac{\pi}{2n}\right) \xrightarrow[]{n\to \infty} 1- 8/\pi^2.
    $$
\end{lemma}
\begin{proof}
Place the vertices at $x_k = r\,e^{2\pi i k/n}$ for $k = 0, \ldots, n-1$.
The chord length between vertices $k$ and $k+j$ is
$\norm{x_k - x_{k+j}} = 2r\sin(\pi j/n)$.
For each gap $j = 1, \ldots, n-1$ there are exactly $n$ pairs $(i,\,i+j)$, so the average per-pair stress is
\begin{equation*}
  S_n = \frac{n}{n(n-1)}\sum_{j=1}^{n-1}\bigl(2r\sin(\tfrac{\pi j}{n}) - 1\bigr)^2
  = \frac{1}{n-1}\sum_{j=1}^{n-1}\bigl(2r\sin(\tfrac{\pi j}{n}) - 1\bigr)^2.
\end{equation*}
Expanding and applying the identities $\sum_{j=1}^{n-1}\sin^2(\pi j/n) = n/2$ and
$\sum_{j=1}^{n-1}\sin(\pi j/n) = \cot(\pi/(2n))$, we get
\begin{equation*}
  S_n = \frac{2n}{n-1}\,r^2 - \frac{4\cot(\pi/(2n))}{n-1}\,r + 1.
\end{equation*}
This convex quadratic in $r$ is minimized at $r_n^* = \cot(\pi/(2n))/n$. The corresponding minimal value of $S_n$ is
$$S^*_n = 1-\frac{2}{n(n-1)}\cot^2\left(\frac{\pi}{2n}\right).$$
As $n \to \infty$, $\cot(\pi/(2n)) \sim 2n/\pi$, so $r_n^* \to 2/\pi$ and $S^*_n \to 1- 8/\pi^2$.
\end{proof}

Recall that $\mathcal{P}(\R^2)$ denotes the set of probability measures on $\R^2$. For $\rho \in \mathcal{P}(\R^2)$ and $X, Y \overset{\text{i.i.d.}}{\sim} \rho$ the average stress functional (corresponding to the raw stress divided by $n(n-1)/2$, not to the stress-1) was defined in Eq.~\ref{eq:stress-functional-maintext} as
\begin{equation}
  \label{eq:stress-functional}
  S[\rho]
  =
  \E_\rho\!\bigl[(\norm{X-Y}-1)^2\bigr] =  \iint \big(\|x-y\|-1\big)^2 \rho(x) \rho(y) \, dx \, dy.
\end{equation}

\begin{lemma}[Optimal circle]
    \label{lem:opt_circle}
    For the uniform measure $\mu_r$ on a circle of radius $r$,
    the stress-minimizing radius is $r^* = 2/\pi$, with optimal stress
    $$S[\mu_{2/\pi}] = 1 - 8/\pi^2 \approx 0.1894$$
\end{lemma}
\begin{proof}
This follows by a limit argument from Lemma~\ref{lem:finiteN}, but for the convenience of the reader, we include a direct proof as well. For $X, Y \overset{\mathrm{i.i.d.}}{\sim} \mu_r$, the angle between them 
$\Phi$ is distributed according to $\mathrm{Uniform}[0, 2\pi)$ and $\norm{X - Y} = 2r\sin(\Phi/2)$.
Using that $\int_0^{2\pi}\sin(\phi/2)\,d\phi = 4$ and
$\int_0^{2\pi}\sin^2(\phi/2)\,d\phi = \pi$, we get
\begin{equation*}
  \E\norm{X-Y} = \frac{4r}{\pi},
  \qquad
  \E\norm{X-Y}^2 = 2r^2.
\end{equation*}
Hence $S[\mu_r] = 2r^2 - 8r/\pi + 1$, which is minimized at $r^* = 2/\pi$
with optimal stress $1 - 8/\pi^2$.
This confirms the continuous limit.
\end{proof}

The following theorem was shown independently by~\citet{mattner1990behandlung, mattner1992extremal} and~\citet{buja1994inequalities}.

\begin{theorem}[Limiting MDS embedding, detailed version of Thm.~\ref{thm:main}]
  \label{thm:main_app}
  Up to translation, the unique minimizer of $S[\rho]$ over $\mathcal{P}(\R^2)$ is the Riesz $s=1$ equilibrium measure with density
  \begin{equation}
    \label{eq:rho_star}
    \rho^*(x)
    =
    \frac{2}{\pi^2\,\sqrt{(\pi/4)^2 - \norm{x}^2}} \, \mathbf{1}\left(\|x\| < \frac{\pi}{4}\right).
  \end{equation}
  It achieves optimal stress
  \begin{equation}
    \label{eq:stress_star}
    S^* = 1 - \frac{\pi^2}{12} \approx 0.1775.
  \end{equation}
\end{theorem}

\begin{proof}
We proceed in eight steps.

Since $S[\rho]$ depends only on pairwise distances $\|X - Y\|$, translating $\rho$ does not affect $S[\rho]$, so we can assume $\E[X] = 0$ without loss of generality. We show in Step 1 that $S$ is strictly convex on the set of zero-centered probability distributions $\rho$. As a result, any such $\rho$ satisfying the Karush--Kuhn--Tucker (KKT) conditions is the unique global minimizer; Steps~2--7 then confirm $\rho^*$ is that point. Note that we need KKT conditions because our optimization problem has both equality and inequality constraints: we optimize Eq.~\ref{eq:stress-functional} over $\rho$ that has to be non-negative and integrate to one.

\medskip
\noindent\textbf{Step~1: Strict Convexity of $\mathbf S$.} To show that $S$ is convex, we consider two zero-centered probability measures $\rho \neq \rho'$. Denoting $\eta = \rho' - \rho$, we have
\begin{align*}
S[\rho'] &=
S[\rho + \eta] = \iint \big(\norm{x-y} -1\big)^2 \big(\rho(x) +\eta(x)\big)(\rho(y)+\eta(y) \big)\, dx\,dy \\
&= S[\rho] +\int \left(2\int \big(\norm{x-y} -1\big)^2 \rho(y)\, dy\right) \eta(x) \, dx + \iint \big(\norm{x-y} -1\big)^2 \eta(x) \eta(y) \, dx\, dy \\
&=S[\rho] + \int \frac{\delta S}{\delta \rho} (x) \eta(x) \, dx + S[\eta].
\end{align*}

Strict convexity means that the functional increases faster than a local linear approximation. The second term is linear in $\eta$ (and the functional $\eta \mapsto \int \tfrac{\delta S}{\delta \rho}\eta$ is the local linear approximation at $\rho$), so to demonstrate strict convexity we need to show that $S[\eta] > 0$. We have
\begin{align*}
  S[\eta] &= \iint \big(\|x-y\|-1\big)^2 \,\eta(x)\eta(y)\,dx\,dy \\
   &=  \iint \big(\norm{x}^2 -2x^\top y +\norm{y}^2- 2\|x-y\| +1 \big) \,\eta(x)\eta(y)\,dx\,dy \\
  &\overset{(a)}{=}  \iint \big( -2x^\top y - 2\|x-y\| \big) \,\eta(x)\eta(y)\,dx\,dy \\
  &=-2 \int x^\top \eta(x)\, dx \int y \eta(y)\, dy -2 \iint \|x-y\| \eta(x)\eta(y)\,dx\,dy \\
  &\overset{(b)}{=} -2\iint \|x-y\| \eta(x)\eta(y)\,dx\,dy \\
  &\overset{(c)}{>} 0,
\end{align*}
where we dropped terms not depending on both $x$ and $y$  at (a) because $\int \eta = 0$ and used $\E_\eta(X) = 0$ at (b). Inequality (c) follows because $\eta\neq 0$ with $\int \eta = 0$ and $\|x-y\|$ is a strictly conditionally negative definite kernel~\citep{schoenberg1938metric}.

Thus, $S$ is strictly convex on the set of zero-centered probability distributions.

\noindent\textbf{Step~2: Expand $S[\rho]$.} Expanding the square in~\eqref{eq:stress-functional}, we get
\begin{equation}
  \label{eq:expand}
  S[\rho]
  = \E\norm{X-Y}^2 - 2\E\norm{X-Y} + 1.
\end{equation}

Note that this identity only holds if all terms on the right are finite. We will now show that this is true for any $\rho$ for which $S[\rho] < \infty$, and we are only interested in this case as we are looking for the minimum of $S$.  The pointwise bound
$a^2 \leq 2(a-1)^2 + 2$, valid since $(a-2)^2 \geq 0$, gives
$\E\norm{X-Y}^2 \leq 2S[\rho] + 2 < \infty$.  We denote $\E\norm{X}^2$ as $\sigma^2$. Conditioning on $X$ yields
\begin{equation*}
  \E_Y\!\bigl[\norm{X-Y}^2\bigr]
  = \E_Y\!\bigl[\norm{X}^2 - 2X\!\cdot\!Y + \norm{Y}^2\bigr]
  = \norm{X}^2 + \E\norm{Y}^2
  = \norm{X}^2 + \sigma^2,
\end{equation*}
where we used $\E[Y] = 0$ and $\E\norm{Y}^2 = \sigma^2$.  Taking the outer expectation, we get $\E\norm{X-Y}^2 = 2\sigma^2$. The left part is finite, hence $\sigma^2 < \infty$.

Thus, 
$$
S[\rho] = 2\sigma^2 - 2\E\norm{X-Y} + 1.
$$

\medskip
\noindent\textbf{Step~3: KKT conditions.}
We can now compute the functional derivative of $S[\rho]$ with respect to $\rho$. For 
$$\sigma^2[\rho] = \int \norm{x}^2 \rho(x)\,dx,$$
the functional is linear in $\rho$
(the integrand $\norm{x}^2$ does not depend on $\rho$), so its functional derivative is simply the integrand:
\begin{equation*}
  \frac{\delta \sigma^2}{\delta \rho}(x) = \norm{x}^2.
\end{equation*}
For the interaction term
\begin{equation*}
  \E\norm{X-Y} = \iint \norm{x-y}\,\rho(x)\rho(y)\,dx\,dy,
\end{equation*}
the functional is quadratic in $\rho$ with symmetric kernel $k(x,y) = \norm{x-y}$.
Thus by the product rule we get
\begin{equation*}
  \frac{\delta}{\delta\rho(x)}\bigl(\E\norm{X-Y}\bigr)
  =
  2\int \norm{x-y}\,\rho(y)\,dy.
\end{equation*}
Together we obtain the functional derivative
$$
\frac{\delta S}{\delta\rho} (x) = 2\norm{x}^2 - 4  \int \norm{x-y}\,\rho(y)\,dy.
$$
This agrees with the expression in Step 1 up to the additive constant $2\sigma^2+2$, which is irrelevant since $\int\eta=0$.

At the minimum, the local linear approximation to $S$ must equal zero or some constraint must be active. Taking the expression for the local linear approximation from Step 1, we obtain the condition for the minimum as
$$
\int \frac{\delta S}{\delta\rho} (x) \eta(x) \, dx = 0,
$$
for all $\eta$ such that $\eta(x) = 0$ for $x\notin \supp(\rho)$. In words, at the optimum, moving an infinitesimal bit of mass from any occupied point $x\in \supp(\rho)$ to any other occupied point $x'\in \supp(\rho)$ must leave $S$ unchanged to first order. This is of course satisfied when $\tfrac{\delta S}{\delta\rho}=0$, but not only then. As we are optimizing over the space of probability distributions, the constraint of them integrating to 1 implies that $\int \eta = 0$, so the condition is also satisfied when $\tfrac{\delta S}{\delta\rho}=C$ for any constant $C$. This is effectively the Lagrange multiplier arising from the equality constraint.

Off the support, we cannot decrease $\rho$ further due to the inequality constraint $\rho(x) \geq 0$. So $\eta$ can only place mass there, not take it away. To ensure that placing mass off the support with some perturbation $\eta$ that has $\eta(x) > 0 $ for some $x\notin \supp(\rho)$ 
does not decrease the functional, $\frac{\delta S}{\delta\rho} (x)$ must be at least as large as on the support. 

Taking this together, we arrive at the full set of KKT conditions at a minimizer: 
\begin{align*}
\frac{\delta S}{\delta\rho} (x) &= C \quad \text{ for } x\in \supp(\rho), \\
\frac{\delta S}{\delta\rho} (x) &\geq C \quad \text{ for } x\notin \supp(\rho).
\end{align*}

Rearranging the on-support
condition gives
\begin{equation}
  \label{eq:star}
  2\,\int \norm{x-y}\,\rho(y)\,dy
  =
  \norm{x}^2 - \tfrac{1}{2}C
  \qquad \text{for }x \in \supp(\rho).
  \tag{$\star$}
\end{equation}

\medskip
\noindent\textbf{Step~4: Laplacian trick.}
Apply $\Delta_x$ to both sides of~\eqref{eq:star}.  For the right-hand side,
$\Delta_x\bigl(\norm{x}^2\bigr) = 4$ in $\R^2$.  For the left-hand
side, we use
\begin{equation}
  \label{eq:laplacian_norm}
  \Delta_x\,\norm{x-y} = \frac{1}{\norm{x-y}} \qquad \text{ for }
  x \neq y.
\end{equation}
To see this, note that we can assume $y=0$ by translation. Then for $r= \|x\|$ we have 
$\Delta r = r'' + r'/r = 0 + 1/r$ by the radial Laplacian formula in $\R^2$. 
 Since $1/r$ is locally
integrable in $\R^2$, $\Delta_x\norm{x-y} = 1/\norm{x-y}$ distributionally
with no delta correction at $x=y$.  Commuting $\Delta_x$ with the integral
(justified for $\rho$ with $\sigma^2 < \infty$ by regularizing $\norm{x-y}_\varepsilon =
\sqrt{\norm{x-y}^2 + \varepsilon^2}$, differentiating under the integral for fixed
$\varepsilon > 0$, then passing $\varepsilon \to 0$ by dominated convergence) yields
\begin{equation}
  \label{eq:starstar}
  \int \frac{\rho(y)}{\norm{x-y}}\,dy = 2
  \qquad \text{for }x \in \supp(\rho).
  \tag{$\star\star$}
\end{equation}

\medskip
\noindent\textbf{Step~5: Riesz equilibrium measure.}
Equation~\eqref{eq:starstar} is the condition that the Riesz $s{=}1$ potential
$$V(x) = \int \frac{\rho(y)}{\norm{x-y}}\,dy$$
is constant (and equal to $2$) on $\supp(\rho)$.  This is satisfied by the Riesz $s=1$ equilibrium~
(\citealp{riesz1938integrales};
 \citealp[§ II.3.13, pp. 163--164, and Appendix, pp. 399--400]{landkof1972foundations};
 \citealp[Lem. 1.2]{chafai2022threshold})
\begin{equation}
  \label{eq:riesz_form}
  \rho^*(x) = \frac{2}{\pi^2 \, \sqrt{(\pi/4)^2 - \norm{x}^2}} \, \mathbf{1}\big(\norm{x} < \pi/4\big).
\end{equation}

\noindent\textbf{Step~6: Verifying the on-support KKT condition.}
The minimizer $\rho^*$ is radially symmetric on the disc $B(0, \pi/4)$, hence the functional derivative $\frac{\delta S}{\delta\rho}$ is also radially symmetric. The functional derivative is also harmonic because
\begin{equation*}
  \Delta \frac{\delta S}{\delta\rho} (x) = 8 - 4V(x) = 0.
\end{equation*}
Any radially symmetric harmonic function must be of the form $a\log\norm{x} +b$. If we show that $\frac{\delta S}{\delta\rho}$ is finite at $x=0$, it will follow that it must be constant. We denote  $R=\pi/4$ and compute
\begin{align*}
   \frac{\delta S}{\delta\rho}(0) &=
  -4\int\norm{y}\,\rho^*(y)\,dy
  =
  -4\int_0^R r\cdot\frac{4r}{\pi\sqrt{R^2-r^2}}\,dr\\
  &\overset{(a)}{=}
  -\frac{16R^2}{\pi}\int_0^{\pi/2}\sin^2\!\theta\,d\theta
  =
  -\frac{16R^2}{\pi}\cdot\frac{\pi}{4}
  =
  -\frac{\pi^2}{4},
\end{align*}
where at (a) we substituted $r=R\sin(\theta)$. Note that this implies $C = \frac{\delta S}{\delta\rho}(0) = -\pi^2 / 4$.

\medskip
\noindent\textbf{Step~7: Verifying the off-support KKT condition.}

\textit{Part~A: $V(x) = 2$ on $\partial B(0,\pi/4)$.}
By Step 5, we have $V(x) = 2$ for all $\norm{x} < \pi/4$, and by continuity $V(x) = 2$ on the boundary.

\textit{Part~B: $V(x) \leq 2$ for $\norm{x} > \pi/4$.}
For $\norm{x} > \pi/4$ (so $\norm{x-y} > 0$ for all
$y \in \supp(\rho^*)$), differentiating under the integral and using
$\Delta_x(1/\norm{x-y}) = 1/\norm{x-y}^3$ yields
\begin{equation*}
  \Delta_x V(x) = \int \frac{\rho^*(y)}{\norm{x-y}^3}\,dy > 0.
\end{equation*}
So $V$ is subharmonic on the unbounded domain $\R^2 \setminus \overline{B(0,\pi/4)}$, with $V = 2$ on $\partial B(0,\pi/4)$
and $V(x) \to 0$ as $\norm{x} \to \infty$.
By the maximum principle, $V \leq 2$ on the entire exterior.

\textit{Part~C: Boundary data at $r = \pi/4$.}
Define 
$$g(x) =\frac{\delta S}{\delta\rho} (x) -C =  2\norm{x}^2 - 4\int\norm{x-y}\rho^*(y)\,dy - C.$$
so that the off-support KKT condition is $g(x) \geq 0$ for $\norm{x} \geq \pi/4$.
Since $\rho^* > 0$ on the open disk, the on-support equality~\eqref{eq:star} holds for
all $\norm{x} < \pi/4$, i.e.\ $g \equiv 0$ on $B(0,\pi/4)$.
The function $g$ is radial (since $\rho^*$ is) and globally $C^1$: for
\begin{equation*}
  \nabla g(x) = 4x - 4\!\int \frac{x-y}{\norm{x-y}}\,\rho^*(y)\,dy,
\end{equation*}
the integrand is dominated by the $L^1$ function $\rho^*$ (since
$\big\|(x-y)/\norm{x-y}\big\| = 1$), so dominated convergence yields continuity of
$\nabla g$ on $\R^2$.  Writing $g(r)$ for the radial profile, the interior identity
$g \equiv 0$ on $B(0,\pi/4)$ together with $C^1$ matching across $r = \pi/4$ gives
\begin{equation}
  \label{eq:bdry_data}
  g(\pi/4) = 0, \qquad g'(\pi/4) = 0.
\end{equation}

\textit{Part~D: Radial ODE outside the disk.}
For $\norm{x} > \pi/4$, differentiating under the integral and using the 2D identity
$\Delta_x\norm{x-y} = 1/\norm{x-y}$ (Step~4) gives
\begin{equation*}
  \Delta g(x) = 8 - 4\,V(x) \geq 0,
\end{equation*}
where the inequality is Part~B.  For radial $g$, $\Delta g = g'' + g'/r$, so
\begin{equation}
  \label{eq:radial_ode}
  \bigl(r\,g'(r)\bigr)' = r\,\Delta g(r) \geq 0 \qquad (r > \pi/4).
\end{equation}

\textit{Part~E: Two integrations.}
By~\eqref{eq:radial_ode}, the function $r\,g'(r)$ is non-decreasing on
$(\pi/4,\infty)$.  For any $r > \pi/4$, the fundamental theorem of calculus and the boundary
data~\eqref{eq:bdry_data} give
\begin{equation*}
  r\,g'(r)\geq \pi/4\cdot g'(\pi/4) = 0,
\end{equation*}
hence $g'(r) \geq 0$ for $r \geq \pi/4$.  A second integration from $\pi/4$ to $r$ yields
\begin{equation*}
  g(r) \geq g(\pi/4) = 0 \qquad (r \geq \pi/4),
\end{equation*}
which is the off-support KKT condition.

\medskip
\noindent\textbf{Step~8: Optimal stress.}
We compute the terms $\sigma^2$ and $\E\norm{X-Y}$ in~\eqref{eq:expand}.
First, 
\begin{align*}
  \sigma^2
    &=
  \int \|x\|^2 \rho(x) dx
  \overset{r= \|x\| }{=} \int_0^R r^2 \cdot \frac{2}{\pi^2\sqrt{R^2-r^2}}2\pi r\,dr
  =
  \frac{4}{\pi}\int_0^R  \frac{r^3}{\sqrt{R^2-r^2}}\,dr \\
  &\overset{(i)}{=}
  \frac{4}{\pi}\int_0^{\pi/2}\frac{R^3\sin^3(\theta)}{R\sqrt{1-\sin^2(\theta)}}R\cos(\theta)\,d\theta
  =
  \frac{4R^3}{\pi} \int_0^{\pi/2} \sin^3(\theta)\, d\theta
  =
  \frac{4R^3}{\pi}\cdot\frac{2}{3}
  =
  \frac{8R^3}{3\pi}
  =
  \frac{\pi^2}{24},
\end{align*}
where at (i) we substituted $r=R\sin(\theta)$.
Integrating~\eqref{eq:star} against $\rho^*$ gives
\begin{equation*}
  \E\norm{X-Y}
  =
  \tfrac{1}{2}\sigma^2 - \tfrac{1}{4}C.
\end{equation*} 

Substituting everything into~\eqref{eq:expand} and using $C=-\pi^2 /4$ from Step 6:
\begin{equation*}
  S^*
  = 2\sigma^2 - 2\E\norm{X-Y} + 1
  = 2\sigma^2 - \sigma^2 + \tfrac{1}{2}C + 1
  =
  \frac{\pi^2}{12} - \frac{\pi^2}{24} - \frac{\pi^2}{8}+ 1
  =
  1 - \frac{\pi^2}{12}
  \approx 0.1775.
\end{equation*}
This completes the proof.
\end{proof}


The final version of our proof is essentially a more verbose version of~\citet{mattner1992extremal}'s argument. We include it for the convenience of the reader. While proving this theorem, we were not aware of the work of~\citet{mattner1992extremal} and~\citet{buja1994inequalities}. Both works were pointed out to us by a MathOverflow user (we will give further details in the deanonymized version).

\begin{corollary}[Distribution of point norms under $\rho^*$]
  \label{cor:norms}
  If $X \sim \rho^*$, the radius $\norm{X}$ has as rescaled  Kumaraswamy $(a{=}2,\,b{=}1/2)$ density
  \begin{equation*}
    f_R(r) = \frac{4r}{\pi\sqrt{(\pi/4)^2 - r^2}} \, \mathbf{1}(r< \pi/4).
  \end{equation*}
\end{corollary}
\begin{proof}
  The density $f_R(r) = 2\pi r\cdot\rho^*(r)$ is obtained by integrating $\rho^*$ over the circle of radius $r$. This is the Kumaraswamy density
  $$ abx^{a-1}(1-x^a)^{b-1}$$
  for $a=2$, $b=1/2$ rescaled from $[0, 1)$ to $[0, \pi/4)$.
\end{proof}

\section{From the Continuum Minimizer to Finite Configurations}
\label{app:perturbation}

Theorem~\ref{thm:main} identifies the continuum minimizer $\rho^*$.  We now
quantify how close the empirical distribution of a finite stress-minimizing
configuration is to $\rho^*$, both for exact unit targets and for
noisy targets.  The natural yardstick is the \emph{energy distance}, whose
square is half the second-order remainder already computed in Step~1 of
the proof of Theorem~\ref{thm:main}, $S[\eta]=2D_E^2(\rho^*+\eta,\rho^*)$, see Lemma~\ref{lem:stability}.

\medskip
\noindent\textbf{Notation and standing assumptions.}
Throughout, $X = (x_1,\ldots,x_n) \in \R^{2\times n}$ with $n \geq 2$, we write
$m = n(n-1)$ for the number of ordered off-diagonal pairs.  The finite MDS stress of the Problem
Formulation is
\begin{equation}
  \label{eq:Sn}
  S_n(X) \;=\; \frac{1}{m} \sum_{i \neq j} \bigl(\norm{x_i - x_j} - 1\bigr)^2,
\end{equation}
a sum over ordered pairs $i \neq j$ only: there are no self-distances,
equivalently the diagonal target distances are set to zero.  Given a symmetric
perturbation $\varepsilon_{ij} = \varepsilon_{ji}$ of the unit targets,
$1 + \varepsilon_{ij}$, the perturbed stress and the root-mean-square
perturbation are
\begin{equation}
  \label{eq:Sneps}
  S_n^\varepsilon(X) \;=\; \frac{1}{m} \sum_{i \neq j}
  \bigl(\norm{x_i - x_j} - 1 - \varepsilon_{ij}\bigr)^2,
  \qquad
  \delta_n^2 \;=\; \frac{1}{m} \sum_{i \neq j} \varepsilon_{ij}^2 .
\end{equation}
Since $S_n$ depends only on the pairwise differences $x_i - x_j$, it is invariant
under a common translation of all points, so we restrict to \emph{centered}
configurations, i.e.\ the linear subspace $V = \{X : \sum_i x_i = 0\} \cong \R^{2(n-1)}$.
On $V$ both $S_n$ and $S_n^\varepsilon$ are \emph{continuous} 
and \emph{coercive}, meaning the value tends to $+\infty$ as $\norm{X} \to \infty$ on
$V$.  A continuous coercive function on a finite-dimensional space
attains its minimum, since its sublevel sets are closed (by continuity) and bounded
(by coercivity), hence compact, so the Weierstrass theorem applies.  We fix global
minimizers

\begin{equation}
  \label{eq:minimizers}
  X_n \in \operatorname*{argmin}_X S_n(X),
  \qquad
  X_n^\varepsilon \in \operatorname*{argmin}_X S_n^\varepsilon(X),
  \qquad
  \alpha_n \;=\; S_n(X_n) \;=\; \min_X S_n(X) .
\end{equation}
To a configuration $X$ we associate the \emph{centered empirical measure}
$\mu_X = \tfrac1n \sum_{i=1}^n \delta_{x_i - \bar x}$ with $\bar x = \tfrac1n\sum_i x_i$;
centering changes neither pairwise distances nor the stress.

\subsection{The energy distance and the stability identity}

\begin{definition}[Energy distance]
  \label{def:energy}
  For probability measures $\mu, \nu$ on $\R^2$ with finite first
  moment, the \emph{squared energy distance} is
  \begin{equation}
    \label{eq:energy_dist}
    D_E^2(\mu,\nu)
    \;=\;
    2\iint \norm{x-y}\,d\mu(x)\,d\nu(y)
    - \iint \norm{x-x'}\,d\mu(x)\,d\mu(x')
    - \iint \norm{y-y'}\,d\nu(y)\,d\nu(y').
  \end{equation}
\end{definition}

Because the Euclidean-distance kernel $\norm{x-y}$ is strictly conditionally
negative definite~\citep{schoenberg1938metric}, one has $D_E^2(\mu,\nu) \geq 0$
with equality if and only if $\mu = \nu$. Thus, $D_E$ is a genuine metric on
distributions~\citep{szekely2013energy,sejdinovic2013equivalence}.  The first finite moment is no real
restriction here. Both $\mu_X$ and $\rho^*$ are compactly supported, and more generally any measure of finite stress has finite second moment (Step~2), hence finite first moment.

The next lemma is the quantitative counterpart of the strict convexity
established in Step~1 of the proof of Theorem~\ref{thm:main}. It upgrades
``$S[\eta] > 0$'' to an exact identity with $D_E^2$ as the leading term.
\begin{lemma}[Quadratic stability]
  \label{lem:stability}
  Let 
  $$g(x)=2\|x\|^2-4\int\|x-y\|\,d\rho^*(y)+\frac{\pi^2}{4}$$
  be the KKT
  slack from the proof of Theorem~\ref{thm:main} (Step~7), so that
  $g \equiv 0$ on $B(0,\pi/4)$ and $g \geq 0$ on $\R^2$.  Then for every centered
  probability measure $\mu$ with finite second moment (equivalently $S[\mu] < \infty$,
  cf.\ Step~2),
  \begin{equation}
    \label{eq:stability_identity}
    S[\mu] - S^*
    \;=\;
    2D_E^2(\mu,\rho^*) \;+\; \int g\,d\mu
    \;\geq\;
   2D_E^2(\mu,\rho^*),
  \end{equation}
  with equality whenever
  $\mu$ is supported in the closed disk $\overline{B}(0,\pi/4)$.
\end{lemma}

\begin{proof}
  Write $\eta = \mu - \rho^*$.  Both $\mu$ and $\rho^*$ are centered probability
  measures, so
  \begin{equation}
    \label{eq:eta_moments}
    \int d\eta = 0, \qquad \int x\,d\eta(x) = 0 .
  \end{equation}
  Since $S$ in~\eqref{eq:stress-functional} is a quadratic form, write
  $$B(\alpha,\beta) = \iint K(x,y)\,d\alpha(x)\,d\beta(y)$$ for the symmetric bilinear
  form with kernel $K(x,y) = (\norm{x-y}-1)^2$, so that $S[\rho] = B(\rho,\rho)$.
  Expanding $S[\rho^* + \eta]$ and using the symmetry $B(\rho^*,\eta) = B(\eta,\rho^*)$,
  \begin{gather}
    \label{eq:bilinear}
    S[\mu]
    \;=\;  B(\rho^*,\rho^*) + 2B(\rho^*,\eta) + B(\eta,\eta)
    \;=\; S[\rho^*] + 2B(\rho^*,\eta) + B(\eta,\eta).
  \end{gather}
  To evaluate the linear coefficient $B(\rho^*,\eta)$, integrate the kernel against
  $\rho^*$ first.  Using $\int x\,d\rho^*(x) = 0$ and $\int\norm{x}^2\,d\rho^*(x) =
  \sigma_*^2$ (with $\sigma_*^2 = \pi^2/24$ from Step~8),
  \begin{align*}
    \int K(x,y)\,d\rho^*(x)
    &\;=\; \int\bigl(\norm{x-y}^2 - 2\norm{x-y} + 1\bigr)\,d\rho^*(x) \\
    &\;=\; \bigl(\norm{y}^2 + \sigma_*^2\bigr) - 2\!\int\norm{x-y}\,d\rho^*(x) + 1 \\
     &\;=\; \underbrace{\norm{y}^2 - 2\!\int\norm{x-y}\,d\rho^*(x)}_{=\,g(y)/2 - \pi^2/8}
     + (\sigma_*^2 + 1),
  \end{align*}
  where the underbraced term is half the first variation $\delta S/\delta\rho$ from Step~3
  evaluated at $\rho^*$, equal to $g(y)/2 - \pi^2/8$ by the definition of $g$.  Therefore
  $$2B(\rho^*,\eta) = \int\bigl(g(y) - \tfrac{\pi^2}{4}\bigr)\,d\eta(y) +
  2(\sigma_*^2 + 1)\int d\eta(y),$$ and since $\int d\eta = 0$ the
  configuration-independent constant $2(\sigma_*^2 + 1)$ drops out, so~\eqref{eq:bilinear}
  becomes
  \begin{equation}
    \label{eq:quad_expand}
    S[\mu] \;=\; S[\rho^*] + \int\Bigl(g(x) - \tfrac{\pi^2}{4}\Bigr)\,d\eta(x) + B(\eta,\eta).
  \end{equation}

  \emph{Linear term.}\;
  Since $\int d\eta = 0$, the constant $\pi^2/4$ in~\eqref{eq:quad_expand} integrates
  to zero, leaving $\int\bigl(g - \tfrac{\pi^2}{4}\bigr)\,d\eta = \int g\,d\eta$.
  Because $g \equiv 0$ on $\supp \rho^*$ we have $\int g\,d\rho^* = 0$, hence
  $\int g\,d\eta = \int g\,d\mu$; and $\int g\,d\mu \geq 0$ because $g \geq 0$
  everywhere (Step~7) and $\mu$ is a nonnegative measure. 

  \emph{Quadratic term.}\;
  Expand $K(x,y) = \norm{x-y}^2 - 2\norm{x-y} + 1$.  The identity
  $\norm{x-y}^2 = \norm{x}^2 - 2x^\top y + \norm{y}^2$ together with the moment
  conditions~\eqref{eq:eta_moments} makes the $\norm{x-y}^2$ and the constant
  $1$ integrate to zero against $\eta \otimes \eta$ (exactly as at steps
  (a)--(b) of Step~1), leaving
  \begin{equation*}
    B(\eta,\eta)
    \;=\;
    -2\iint \norm{x-y}\,d\eta(x)\,d\eta(y)
    \;=\;
    2D_E^2(\mu,\rho^*),
  \end{equation*}
  where the second equality is the bilinear expansion of $\eta = \mu - \rho^*$,
  matching~\eqref{eq:energy_dist}.
 
  Substituting both terms
  into~\eqref{eq:quad_expand} yields the identity~\eqref{eq:stability_identity};
  dropping $\int g\,d\mu \geq 0$ gives the inequality.  If $\supp\mu \subseteq
  \overline B(0,\pi/4)$ then $\int g\,d\mu = 0$ and equality holds.
\end{proof}

\subsection{Exact targets: finite minimizers converge to $\rho^*$}

We first record how the finite stress~\eqref{eq:Sn} sits inside the continuum
functional~\eqref{eq:stress-functional}.  Evaluating $S$ at the empirical measure counts
all $n^2$ ordered pairs, and the $n$ diagonal pairs each contribute
$(\norm{x_i - x_i} - 1)^2 = 1$:
\begin{equation}
  \label{eq:finite_continuum}
  S[\mu_X]
  \;=\;
  \frac{1}{n^2}\sum_{i,j}\bigl(\norm{x_i - x_j} - 1\bigr)^2
  \;=\;
  \frac{1}{n^2}\Bigl[\,m\,S_n(X) + n\,\Bigr]
  \;=\;
  \frac{n-1}{n}\,S_n(X) + \frac{1}{n}.
\end{equation}
The additive $\tfrac{1}{n}$ is the self-pair contribution. With the diagonal
targets set to zero it is absent from $S_n$, but the atomic measure $\mu_X$
places mass $1/n^2$ on each diagonal pair of $\mu_X \otimes \mu_X$.  Crucially
this term is independent of $X$, so it neither affects which configuration
minimizes nor survives in the leading order of the bounds below.  Since the scalar
map $t \mapsto \tfrac{n-1}{n}t + \tfrac{1}{n}$ is strictly increasing,
\eqref{eq:finite_continuum} gives $S[\mu_X] \leq S[\mu_{X'}]$ if and only if
$S_n(X) \leq S_n(X')$ for any two configurations $X, X'$; hence $S_n$ and
$X \mapsto S[\mu_X]$ share the same minimizers, and in particular the finite
minimizer $X_n$ of $S_n$ also minimizes $X \mapsto S[\mu_X]$ over all $n$-point configurations.

\begin{proposition}[Finite minimizers under exact targets]
  \label{prop:finite_exact}
  For $a_n = S[\mu_{X_n}] - S^* \geq 0$ we have
  \begin{equation}
    \label{eq:an_bound}
    0 \;\leq\; a_n \;\leq\; \frac{1 - S^*}{n} \;=\; \frac{\pi^2}{12n},
  \end{equation}
  and hence the empirical measure $\mu_{X_n}$ of the finite minimizer deviates from the continuum minimizer $\rho^*$ by
  \begin{equation}
    \label{eq:DE_exact}
    D_E(\mu_{X_n},\rho^*) \;\leq\; \frac{\pi}{\sqrt{24\,n}}.
  \end{equation}
  Moreover, the minimal off-diagonal stress is squeezed against the continuum optimizer,
  \begin{equation}
    \label{eq:alpha_bound}
    1 - \frac{n}{n-1} \frac{\pi^2}{12} =
    S^* - \frac{1 - S^*}{n-1}
    \;\leq\; \alpha_n \;\leq\;
    S^* = 1 - \frac{\pi^2}{12}.
  \end{equation}
  
\end{proposition}

\begin{proof}
  Let $Y_1,\ldots,Y_n \overset{\mathrm{i.i.d.}}{\sim} \rho^*$.  Since the
  summands of $S_n(X)$~\eqref{eq:Sn} are identically distributed with
  $$\E(\norm{Y_i - Y_j} - 1)^2 = S[\rho^*] = S^* \quad \text{ for } i \neq j,$$
  we have
  \begin{equation}
    \label{eq:ESn}
    \E\,S_n(Y) \;=\; S^*.
  \end{equation}
  As a result, there must be some configuration that does not exceed this expectation, so that 
  \begin{equation}
    \alpha_n \;=\; \min_X S_n(X) \;\leq\; \E\,S_n(Y) \;=\; S^*,
  \end{equation}
  which is the upper bound in~\eqref{eq:alpha_bound}.  Applying the affine
  relation~\eqref{eq:finite_continuum} and using that $X_n$ minimizes
  $S[\mu_X]$,
  \begin{equation*}
    S[\mu_{X_n}] \;=\; \min_X S[\mu_X]
    \;\leq\; \E\,S[\mu_Y]
    \;=\; \frac{n-1}{n}\,\E S_n(Y) + \frac{1}{n}
    \;=\; \frac{n-1}{n}\,S^* + \frac{1}{n}.
  \end{equation*}
  Subtracting $S^*$ gives 
  $$a_n \leq \frac{1-S^*}{n} = \frac{\pi^2}{12n},$$
  and $a_n \geq 0$ because $\rho^*$ is the global
  minimizer of $S$ over all probability measures (Theorem~\ref{thm:main}),
  so $S[\mu_{X_n}] \geq S^*$.  This is~\eqref{eq:an_bound}.   Lemma~\ref{lem:stability} then yields
  $$2D_E^2(\mu_{X_n},\rho^*) \leq S[\mu_{X_n}] - S^* = a_n\leq  \frac{\pi^2}{12n},$$
    which is~\eqref{eq:DE_exact}.  Finally, $a_n \geq 0$ combined
  with~\eqref{eq:finite_continuum} reads
  $$\frac{n-1}{n}\alpha_n + \frac{1}{n} \geq S^*,$$
  which we can rearrange to the lower bound in~\eqref{eq:alpha_bound}.
\end{proof}

\subsection{Noisy targets: a deterministic perturbation bound}

The passage from exact to noisy targets rests on a single deterministic
inequality, the reverse triangle inequality in the space of pairwise residuals.

The statement holds for arbitrary target distances, so we phrase it that way.  For a
symmetric off-diagonal dissimilarity matrix $D = (d_{ij})_{i\neq j}$ write
\begin{equation*}
  S_D(X) \;=\; \frac{1}{m}\sum_{i\neq j}\bigl(\norm{x_i - x_j} - d_{ij}\bigr)^2,
  \qquad
  \norm{D - D'} \;=\; \Bigl(\frac{1}{m}\sum_{i\neq j}(d_{ij} - d'_{ij})^2\Bigr)^{1/2},
\end{equation*}
so that $S_n = S_{\mathbf 1}$ and $S_n^\varepsilon =S_{\mathbf 1 + \varepsilon}$
with $\norm{(\mathbf 1 + \varepsilon) - \mathbf 1} = \delta_n$.

\begin{lemma}[Reverse triangle inequality for the stress]
  \label{lem:revtri}
  Let $D, D'$ be dissimilarity matrices.
  \begin{enumerate}
  \item[\textit{(i)}] For every configuration $X$,
    \begin{equation}
      \label{eq:revtri}
      \Bigl| \sqrt{S_D(X)} - \sqrt{S_{D'}(X)} \Bigr| \;\leq\; \norm{D - D'} .
    \end{equation}
  \item[\textit{(ii)}] If $X_D$ and $X_{D'}$ minimize $S_D$ and $S_{D'}$,
    then
    \begin{equation}
      \label{eq:perturb_stress}
      S_D(X_{D'}) - S_D(X_D)
      \;\leq\;
      4\,\norm{D - D'}\sqrt{S_D(X_D)} + 4\,\norm{D - D'}^2 .
    \end{equation}
  \end{enumerate}
  
\end{lemma}
  The square root of a raw stress is the Euclidean norm of its residual vector~\citep{borg2005modern}, which makes \textit{(i)} an instance of the reverse triangle
  inequality.  Related perturbation bounds for stress minimization under noise, under
  stronger assumptions and by a different argument, appear in~\citep{ariascastro2023stability}.
\begin{proof}
  Fix $X$ and collect the residuals into vectors in $\R^m$,
  \begin{equation*}
    a = \tfrac{1}{\sqrt m}\bigl(\norm{x_i - x_j} - d_{ij}\bigr)_{i\neq j},
    \qquad
    a' = \tfrac{1}{\sqrt m}\bigl(\norm{x_i - x_j} - d'_{ij}\bigr)_{i\neq j},
  \end{equation*}
  so that 
  $$\norm{a}_2 = \sqrt{S_D(X)}, \quad \norm{a'}_2 = \sqrt{S_{D'}(X)},$$
  and 
  $$\norm{a - a'}_2 = \norm{D - D'}.$$
  The triangle inequality gives $\norm{a}_2 \leq \norm{a'}_2 + \norm{a - a'}_2$ and,
  interchanging $a$ and $a'$, $\norm{a'}_2 \leq \norm{a}_2 + \norm{a - a'}_2$, hence
  $$\bigl|\,\norm{a}_2 - \norm{a'}_2\,\bigr| \leq \norm{a - a'}_2 = \norm{D - D'},$$
   which
  is~\eqref{eq:revtri}.

  For \textit{(ii)}, apply~\eqref{eq:revtri} at $X_{D'}$, then optimality of $X_{D'}$ for
  $S_{D'}$, then~\eqref{eq:revtri} at $X_D$,
  \begin{equation*}
    \sqrt{S_D(X_{D'})}
    \;\leq\; \sqrt{S_{D'}(X_{D'})} + \norm{D - D'}
    \;\leq\; \sqrt{S_{D'}(X_D)} + \norm{D - D'}
    \;\leq\; \sqrt{S_D(X_D)} + 2\,\norm{D - D'} .
  \end{equation*}
  The term $2\norm{D - D'}$ arises because the perturbation is used twice.  Squaring
  gives~\eqref{eq:perturb_stress}.
\end{proof}

Taking $D = \mathbf 1$ and $D' = \mathbf 1 + \varepsilon$, so that $\norm{D - D'} = \delta_n$,
$X_D = X_n$, $X_{D'} = X_n^\varepsilon$, and $\alpha_n = S_n(X_n)$ as in~\eqref{eq:minimizers},
inequality~\eqref{eq:perturb_stress} reads
\begin{equation}
  \label{eq:perturb_stress_unit}
  S_n(X_n^\varepsilon) - \alpha_n
  \;\leq\;
  4\,\delta_n\sqrt{\alpha_n} + 4\,\delta_n^2 .
\end{equation}

\begin{corollary}[Finite minimizers under noisy targets]
  \label{cor:finite_noisy}
  The minimizer $X_n^\varepsilon$ of the perturbed stress satisfies
  \begin{equation}
    \label{eq:DE_noisy}
    D_E^2(\mu_{X_n^\varepsilon},\rho^*)
    \;\leq\;
    \frac{\pi^2}{24n}
    \;+\; 2\sqrt{S^*}\,\delta_n
    \;+\; 2\,\delta_n^2
    \;\approx\;
    \frac{0.411}{n} + 0.843\,\delta_n + 2\,\delta_n^2 .
  \end{equation}
  In particular, perturbed finite minimizers remain close to $\rho^*$ whenever
  both $n^{-1}$ and the RMS perturbation $\delta_n$ are small; the worst-case
  dependence on the noise is $D_E = O(\sqrt{\delta_n})$.
\end{corollary}

\begin{proof}
  By Lemma~\ref{lem:stability}, adding and subtracting the exact-target value
  $S[\mu_{X_n}]$,
  \begin{equation*}
    2D_E^2(\mu_{X_n^\varepsilon},\rho^*)
    \;\leq\; S[\mu_{X_n^\varepsilon}] - S^*
    \;=\; \bigl(S[\mu_{X_n^\varepsilon}] - S[\mu_{X_n}]\bigr) + \bigl(S[\mu_{X_n}] - S^*\bigr).
  \end{equation*}
  The second bracket is $S[\mu_{X_n}] - S^* = a_n \leq \tfrac{\pi^2}{12n}$
  by~\eqref{eq:an_bound}.  In the first bracket, we use the affine
  relation~\eqref{eq:finite_continuum},
  \begin{equation*}
    S[\mu_{X_n^\varepsilon}] - S[\mu_{X_n}]
    \;=\; \frac{n-1}{n}\bigl(S_n(X_n^\varepsilon) - S_n(X_n)\bigr)
    \;=\; \frac{n-1}{n}\bigl(S_n(X_n^\varepsilon) - \alpha_n\bigr)
    \;\leq\; \frac{n-1}{n}\bigl(4\delta_n\sqrt{\alpha_n} + 4\delta_n^2\bigr),
  \end{equation*}
  using~\eqref{eq:perturb_stress_unit}.  Bounding $\tfrac{n-1}{n} \leq 1$,
  $\sqrt{\alpha_n} \leq \sqrt{S^*}$ from~\eqref{eq:alpha_bound}, and dividing by $2$ gives~\eqref{eq:DE_noisy}, with numerical constants from $\pi^2/24 \approx 0.411$ and
  $2\sqrt{S^*} = 2\sqrt{1 - \pi^2/12} \approx 0.843$.
\end{proof}

\begin{remark}[Evaluating $D_E$ numerically]
  \label{rem:numerics}
  The quantity most aligned with the theory is $D_E(\mu_n,\rho^*)$, computable
  from~\eqref{eq:energy_dist} as
  \begin{equation}
    \label{eq:DE_estimator}
    D_E^2(\mu_n,\rho^*)
    \;=\;
    \frac{2}{n}\sum_{i=1}^n \E_{Y\sim\rho^*}\norm{x_i - Y}
    \;-\; \frac{1}{n^2}\sum_{i,j}\norm{x_i - x_j}
    \;-\; \E_{Y,Y'\sim\rho^*}\norm{Y - Y'} .
  \end{equation}
  The self-interaction constant is
  $\E_{Y,Y'\sim\rho^*}\norm{Y-Y'} = \tfrac{\pi^2}{12} \approx 0.822$ (Step~8).
  For a point inside the support, $\norm{x} \leq \pi/4$, the on-support
  identity~\eqref{eq:star} gives the closed form
  $\E_{Y\sim\rho^*}\norm{x - Y}
  = \tfrac12\bigl(\norm{x}^2 + \tfrac{\pi^2}{8}\bigr)$
\end{remark}

\begin{remark}
    Note that if the mean $\bar{\varepsilon}$ of the $\varepsilon_{ij}$ over $ij$ is zero, this $\delta_n$ is equal to the standard deviation of the perturbation values. If the mean is not zero, our bound still holds. But a potentially better bound could be obtained by attributing $\bar{\varepsilon}$ of $\varepsilon_{ij}$ to the equidistant part of the target distances and rescaling $\rho^*$ and the constants in the bound by $1+\bar{\varepsilon}$.
\end{remark}

\end{document}

%% file: math_commands.tex
\usepackage{amsmath,amsfonts,bm}

\def\eqref#1{equation~\ref{#1}}

\def\1{\bm{1}}

\DeclareMathAlphabet{\mathsfit}{\encodingdefault}{\sfdefault}{m}{sl}
\SetMathAlphabet{\mathsfit}{bold}{\encodingdefault}{\sfdefault}{bx}{n}

\newcommand{\E}{\mathbb{E}}

\newcommand{\R}{\mathbb{R}}

